\pdfoutput=1
\documentclass[11pt]{article}
\usepackage[final]{acl}
\usepackage{times}
\usepackage{latexsym}
\usepackage[T1]{fontenc}
\usepackage[utf8]{inputenc}
\usepackage{microtype}
\usepackage{inconsolata}
\usepackage{graphicx}
\usepackage{subfigure}
\usepackage{tcolorbox}
\usepackage{colortbl}
\usepackage{hyperref}
\usepackage{float}

\usepackage{pifont}       
\usepackage{bbding}       
\usepackage{fontawesome}  

\usepackage{booktabs}
\usepackage{multirow}  
\usepackage{multicol}  

\title{MMXU: A Multi-Modal and Multi-X-ray Understanding Dataset for Disease Progression}

\author{
 \textbf{Linjie Mu\textsuperscript{1,2}},
 \textbf{Zhongzhen Huang\textsuperscript{1,2}},
 \textbf{Shengqian Qin\textsuperscript{1,2}},
 \textbf{Yakun Zhu\textsuperscript{1,2}},
\\
 \textbf{Shaoting Zhang\textsuperscript{1}},
 \textbf{Xiaofan Zhang\textsuperscript{1,2,3\thanks{~~Corresponding author}}}
\\
\\
 \textsuperscript{1}Shanghai Jiao Tong University,
 \textsuperscript{2}SPIRAL Lab,
 \textsuperscript{3}SII
\\
 \small{
   \textbf{Correspondence:} 
   \href{xiaofan.zhang@sjtu.edu.cn}{xiaofan.zhang@sjtu.edu.cn}
 }
}

\begin{document}
\maketitle
\begin{abstract}

Large vision-language models (LVLMs) have shown great promise in medical applications, particularly in visual question answering (MedVQA) and diagnosis from medical images. However, existing datasets and models often fail to consider critical aspects of medical diagnostics, such as the integration of historical records and the analysis of disease progression over time. In this paper, we introduce MMXU (Multimodal and MultiX-ray Understanding), a novel dataset for MedVQA that focuses on identifying changes in specific regions between two patient visits. Unlike previous datasets that primarily address single-image questions, MMXU enables multi-image questions, incorporating both current and historical patient data. We demonstrate the limitations of current LVLMs in identifying disease progression on MMXU-\textit{test}, even those that perform well on traditional benchmarks. To address this, we propose a MedRecord-Augmented Generation (MAG) approach, incorporating both global and regional historical records.
Our experiments show that integrating historical records significantly enhances diagnostic accuracy by at least 20\%, bridging the gap between current LVLMs and human expert performance. Additionally, we fine-tune models with MAG on MMXU-\textit{dev}, which demonstrates notable improvements. We hope this work could illuminate the avenue of advancing the use of LVLMs in medical diagnostics by emphasizing the importance of historical context in interpreting medical images.
Our dataset is released at github\footnote{\quad\url{https://github.com/linjiemu/MMXU}}.

\end{abstract}

\section{Introduction}

\begin{figure}[t]
  \includegraphics[width=\columnwidth]{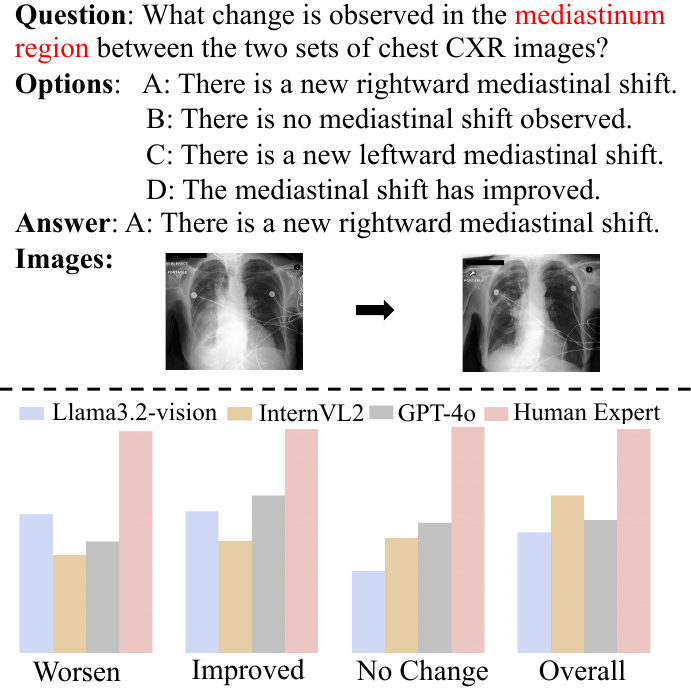}
  \vspace{10pt}
  \caption{The upper part of the figure presents an example from our constructed dataset, which includes two images and a question regarding their changes. The lower part illustrates the performance of human experts and several models on three types of disease progression questions, emphasizing a notable gap between the models and human experts.}
  \label{fig:dataset}
\end{figure}

Stemming from the ever-increasing number of parameters and large-scale training corpus, large vision-language models (LVLMs)~\cite{zhu2023minigpt,bai2023qwen,liu2023llava,achiam2023gpt,lu2024deepseek,chen2024internvl} have demonstrated remarkable capabilities in general visual scene perception and understanding, as well as in generating textual descriptions. As the development of LVLMs accelerates, this paradigm has spurred significant advancements in the medical field~\cite{li2023chatdoctor,wu2024pmc,li2024llava,chen2024huatuogpto1medicalcomplexreasoning}, particularly in the analysis and diagnosis of medical images.

Although current medical LVLMs have shown strong performance in downstream tasks such as medical visual question answering (MedVQA)~\cite{hu2024omnimedvqa,sun-etal-2024-self,saeed-2024-medifact} and medical report generation~\cite{zhou-wang-2024-divide,bu-etal-2024-dynamic,huang-etal-2025-cmeaa,yin-etal-2025-kia} on public benchmarks, they are still limited in responding to basic visual questions (i.e., those involving a single image and a brief description ~\cite{bae2024ehrxqa,liu2024gemex}). In real-world scenarios, diagnosis often requires physicians to integrate both case history and current evidence~\cite{lorkowski2022medical}. Case history serves as an invaluable source of evidence, encompassing factors such as previous medical conditions, treatments, and patient demographics~\cite{liu2024gemex}. Identifying the differences between symptoms, clinical signs, and diagnostic outcomes across the timeline is key to accurate diagnosis. Based on these considerations, we ask: Can LVLMs enhance the diagnostic process by identifying these critical factors?

In this paper, we introduce MMXU (Multimodal and MultiX-ray Understanding), leveraging the rich resources of patient electronic medical records (EMRs) from the MIMIC-CXR dataset. As shown in Figure \ref{fig:dataset}, unlike previous works~\cite{bae2024ehrxqa,liu2024gemex}, which focus solely on the current image or are restricted to simple questions, this benchmark is specifically designed to inquire about differences in specific regions between a patient’s two visits. MMXU contains two splits: \textit{test} and \textit{dev}. MMXU-\textit{test} consists of 3,000 entries from 1,201 patients and 2,469 studies, while MMXU-\textit{dev} contains 118K QA pairs involving 114K images. We evaluate a range of open-source and closed-source models and conduct a user study to assess human performance. The results indicate that LVLMs struggle to identify differences between two visits, even the LVLM that demonstrates notable performance on other medical benchmarks exhibits a significant performance gap—nearly 40\%—when compared to human performance.


Recognizing the limitations of current LVLMs in such scenarios, we take steps toward bridging this gap. 
Since physicians often rely on patient history, previous treatments, and other contextual information during the diagnostic process, we propose a novel approach, MedRecord-Augmented Generation (MAG) to facilitate LVLMs. We introduce two types of records as contextual information for diagnosis: global records (e.g., patient demographics and overall medical report) and regional records (e.g., specific regional details and diagnostic results). 
Our experiments show that incorporating global or regional medical records leads to a significant accuracy improvement of at least 20\% for several well-known models.
Acknowledging the inadequate training on such tasks, 
we apply the MAG method for fine-tuning the MMXU-\textit{dev} dataset and substantiate the effectiveness of both the dataset and the MAG approach in enhancing overall performance.

In summary, this paper presents three key contributions. First, we introduce MMXU, the first multi-image MedVQA dataset designed to investigate differences in specific regions of CXR images between a patient’s two visits, addressing the gap between current MedVQA benchmarks and real-world clinical scenarios. Second, our evaluation results show that current large vision-language models (LVLMs), including proprietary models such as GPT-4o, face significant challenges in identifying disease progression between two visits. Third, we propose the MedRecord-Augmented Generation (MAG) method to enhance the diagnosis of X-rays over time by leveraging the patient’s global and regional historical records. Experimental results demonstrate the effectiveness of MAG, underscoring the potential of incorporating contextual information in diagnostic processes.

\begin{table*}[ht]
    \centering\small
    \resizebox{\textwidth}{!}{
        \begin{tabular}{ccccccc}
            \toprule
            \textbf{Dataset} & \textbf{\#Images} & \textbf{\# QA Pairs} & \textbf{\#Question Type} & \textbf{\#Regional} & \textbf{\#Comparative} &\textbf{\#Complex}\\
            \midrule
            \textbf{MIMIC-CXR-VQA}~\cite{bae2024ehrxqa}&142K&377K&Single&\ding{51}&\ding{55}&\ding{51}                  \\
            \textbf{GEMeX}~\cite{liu2024gemex}&151K&1.6M&Single&\ding{51}&\ding{55}&\ding{51}                           \\
            \textbf{Medical-Diff-VQA}~\cite{hu2023medical}&164K&700K&Single\&Multiple&\ding{55}&\ding{51}&\ding{55}       \\
            \textbf{MMXU-\textit{dev} (Ours)}&114K&118K&Multiple&\ding{51}&\ding{51}&\ding{51}                                        \\
            \bottomrule
        \end{tabular}
    }
    \vspace{10pt}
    \caption{Comparison of the MedVQA dataset constructed on MIMIC-CXR. In the \#Question Type column, ``Single'' refers to questions about a single image, and ``Multiple'' refers to questions about multiple images. Most existing datasets primarily focus on observations in a single image. While Medical-Diff-VQA contains 131k QA pairs that focus on multi-image changes, it has a simple structure and only considers global-level differences. Our MMXU-\textit{dev} dataset is the first to focus on complex changes in the same regions across multiple images at the regional level.}
    \label{tab:dataset-comparison}
    
\end{table*}

\section{Related Work}

\noindent \textbf{Large Vision-Language Models} ~ ~ Large Vision-Language Models, which integrate vision encoders, connectors, and large language models to enhance cross-modal understanding, have emerged as powerful frameworks that combine visual and textual information for a wide range of tasks. These models can be systematically categorized based on the type of connector. The first category comprises approaches utilizing cross-attention-based connectors, such as Flamingo~\cite{alayrac2022flamingo} and CogVLM~\cite{wang2023cogvlm,hong2024cogvlm2}, which exploit attention mechanisms to facilitate the exchange of information between the vision encoder and the language model. The second category includes methods that employ query-based connectors, such as BLIP-2~\cite{li2023blip}, Instruct-BLIP~\cite{dai2023instructblip}, mPLUG-owl2~\cite{ye2024mplug}, and Qwen-VL~\cite{bai2023qwen}, wherein queries are leveraged to orchestrate the interaction between visual and textual modalities, thereby enhancing the alignment and coherence of visual and linguistic representations. Furthermore, projection-based connector methods, exemplified by LLaVA~\cite{liu2023llava}, Mini-GPT4~\cite{zhu2023minigpt}, DeepSeek-VL~\cite{lu2024deepseek}, and Mini-Gemini~\cite{li2024mini}, project visual data into a shared embedding space, thereby fostering seamless integration with textual information.
These innovations offer a range of solutions for cross-modal understanding, driving the potential applications of intelligent systems in multi-task learning.


\noindent \textbf{MedVQA Dataset on radiology} ~ ~ Medical visual question answering (MedVQA) datasets play a pivotal role in advancing AI-driven clinical decision-making. VQA-RAD~\cite{Lau2018}, as an early pioneering work, introduces a meticulously curated dataset for radiology images, featuring clinician-generated questions and corresponding answers tailored to clinically relevant tasks. SLAKE~\cite{liu2021slake} stands out as a large, bilingual dataset, enriched with extensive semantic annotations and spanning a wide range of radiological modalities. 

MIMIC-CXR~\cite{johnson2019mimic} provides a vast collection of 371,920 chest X-rays from 65,079 patients, serving as the foundation for numerous subsequent studies. The comparison of these datasets is shown in Table \ref{tab:dataset-comparison}.  MIMIC-CXR-VQA~\cite{bae2024ehrxqa} seamlessly integrates chest X-rays with Electronic Health Records (EHRs), facilitating multi-modal question answering with an emphasis on region-specific queries. The Medical-Diff-VQA~\cite{hu2023medical} is notable for its inclusion of seven distinct question types, particularly focusing on the comparative analysis of current and reference images for diagnostic purposes. GEMeX~\cite{liu2024gemex} offers a large-scale, explainable VQA benchmark, complete with detailed visual and textual explanations, thus addressing the growing need for a diverse array of clinical questions. Most of these datasets primarily focus on observations from a single image. While Medical-Diff-VQA contains 131k QA pairs that address multi-image changes, its structure is relatively simple and only accounts for global-level differences. Our MMXU-\textit{dev} dataset is the first to understand complex changes in the same regions across multiple chest X-ray images of the same patient at the regional level, spanning several visits.

\section{Dataset Construction}
\begin{figure*}[t]
  \centering
  \includegraphics[width=2.05\columnwidth]{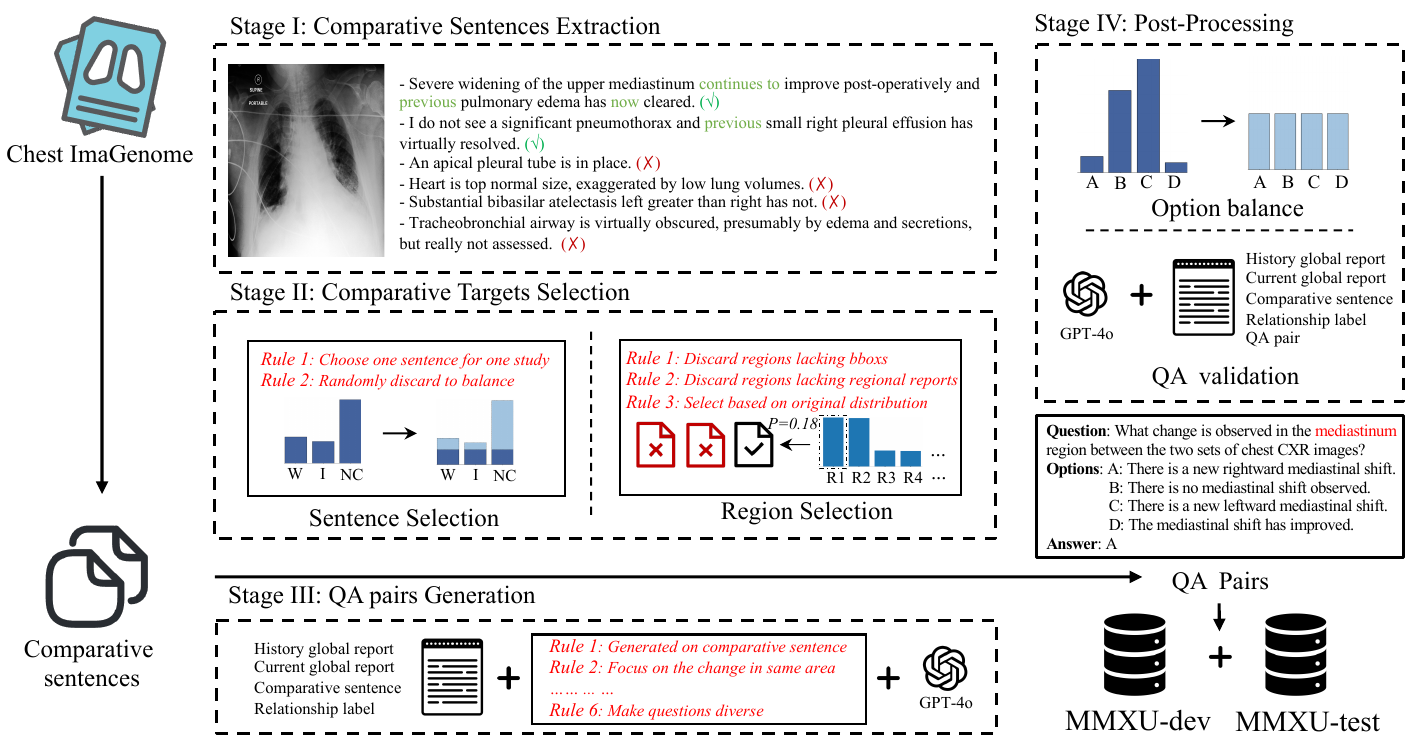}
  \vspace{10pt}
  \caption{The pipeline of constructing MMXU. In the first stage, we extract all comparative sentences and their associated labels from the Chest ImaGenome dataset. In the second stage, we refine our selection by applying precise filtering rules and identifying regions of interest within these sentences. The third stage involves leveraging GPT-4o with meticulously crafted prompts to generate region-level question-answer pairs. In the final stage, we further refine and filter generated QA pairs, thereby constructing the MMXU-\textit{test} and MMXU-\textit{dev}.}
  \label{fig:framework}
\end{figure*}

In this section, we outline the pipeline of constructing MMXU, as shown in Figure \ref{fig:framework}. The process starts with the Chest ImaGenome dataset~\cite{wu2021chest}, which includes the \textit{silver\_dataset} section containing annotations for 243,310 images from 63,945 patients. These annotations cover bounding boxes for 29 anatomical regions, along with corresponding region-level report \textit{phrases}, labeled \textit{attributes}, and \textit{relationships}. The entire method consists of four distinct phases: (1) Comparative Sentences Extraction (\S \ref{sec:stage1}), (2) Comparative Targets Selection (\S \ref{sec:stage2}), (3) QA pairs Generation (\S \ref{sec:stage3}), and (4) Post-Processing (\S \ref{sec:stage4}).

\subsection{Comparative Sentences Extraction}
\label{sec:stage1}
In the first stage, our objective is to identify and extract sentences that contain comparative information, forming the foundation for generating question-answer pairs in subsequent stages. For example, the sentence ``Previously seen ill-defined peribronchial lower lobe opacity seen on lateral view has resolved,'' along with its associated \textit{relationships} label ``comparison|yes|improved'' explicitly indicates a comparison with prior conditions, highlighting the resolution of the lower lobe opacity and thus signaling an improvement in the patient's condition. In total, we extract 232,247 comparative sentences, encompassing 22,770 patients and 102,606 reports. A more detailed example of such a sentence is provided in Appendix \ref{app:stage1}.

\subsection{Comparative Targets Selection}
\label{sec:stage2}
At this stage, we categorize comparative sentences into three groups based on disease progression: ``Worsen'' (108,734), ``Improved'' (91,084), and ``No Change'' (264,676), with the majority concentrated in the ``No Change'' category. Given that the prevalence of medical conditions often follows a long-tail distribution~\cite{wu2024medical}, it is essential to achieve a more balanced dataset for a robust evaluation of LVLMs. To this end, we implement two key principles for sentence selection: (1) Retaining only one comparative sentence per report, prioritizing those indicative of change, to enhance data diversity and reduce redundancy. (2) Randomly discarding a portion of ``No Change'' sentences to equalize their count with that of the ``Improved'' and ``Worsen'' categories, thereby mitigating class imbalance.

Next, we implement a regional selection process to improve query accuracy, where one region is chosen for each sentence to question. In real-world applications, a comparative statement may reference multiple anatomical regions simultaneously. For instance, the sentence ``In comparison with the study of \_\_\_, there has been worsening of the increased opacification at the left base with silhouetting of the hemidiaphragm and blunting of the costophrenic angle'' concurrently describes the ``left lung'', ``left lower lung zone'', ``left costophrenic angle'' and ``left hemidiaphragm''. Consequently, it is crucial to select the precise regions mentioned in the sentence to eliminate ambiguity. Our approach is as follows: (1) Exclude regions in both current and historical images that lack a clearly defined bounding box. (2) Eliminate regions that are not referenced in prior reports, as it is infeasible to retrieve historical region-level reports. (3) In cases where multiple regions remain, randomly select one based on the original distribution.

\subsection{QA pairs Generation}
\label{sec:stage3}

In the third stage, we employ GPT-4o as the generator, leveraging comparative sentences, \textit{relationship} labels, and reports from two visits as foundational data to guide the model in producing question-answer pairs. Our observations suggest that a single sentence can encapsulate multiple changes. For instance, the sentence ``Bilateral pleural effusions are again seen, and atelectasis is present'' concurrently describes alterations in both ``pleural effusions'' and ``atelectasis''. To fully harness this data, we direct GPT-4o to generate up to three distinct question-answer pairs, ensuring maximum diversity. Additionally, to facilitate the future validation of data accuracy, we require the model to provide justifications for the generated pairs. 

In total, we have established six rules to guide GPT-4o generating QA pairs that capture changes in the same area across the two reports. 
Appendix \ref{app:stage3} presents our prompt template and a detailed example during the QA generation process.

\begin{figure}[t]
    \centering\small
\begin{tcolorbox}[colframe=black]
\ttfamily
Here are two chest X-RAY images reports of the same patient. Previous chest X-RAY is Image-1. Current chest X-RAY is Image-2. Your task is to evaluate the differences between the two images based on the provided report and question.\\
The report of Image-1 is: \%s
\\
The report of Image-2 is: \%s
\\
Question:\%s
\\
Options:\\
A: \%s\\
B: \%s\\
C: \%s\\
D: \%s\\
Answer with the option's letter from the given choices directly.

\end{tcolorbox}
\caption{Text template during the data post-processing.}
\label{fig:post-processing}
\vspace{10pt}
\end{figure}

\subsection{Post-Processing}
\label{sec:stage4}

Finally, we conduct post-processing to further ensure the quality of generated QA-pairs. Since imbalanced answer options can mislead the model during training and result in unfair evaluations during testing, we first balance the distribution of four answer choices in the QA pairs generated by GPT-4o. Moreover, we employ GPT-4o to answer the generated questions with the prompt detailed in Figure  \ref{fig:post-processing}, which contains the information used during question generation. Questions that were answered incorrectly are considered either excessively difficult or erroneous and are removed. As illustrated in Table \ref{tab:post}, 95.1\% of the questions were answered correctly. The final candidate set comprises 121,800 QA pairs, including 47,000 instances of ``Worsen'', 41,000 instances of ``Improved'' and 35,000 instances of ``No Change''.

\begin{table}[ht]
    \centering\small
    \begin{tabular}{c|ccc}
        \toprule
       & \textbf{\#Examples} & \textbf{\#Correctness} & \textbf{Rate}\\
        \midrule
        \textbf{Worsen}&50,305&45,395&90.2\%\\
        \textbf{Improved}&40,647&39,956&98.3\%\\
        \textbf{No change}&37,124&33,449&90.1\%\\
        \textbf{Overall}&128,076&121,800 & 95.1\%\\
        \bottomrule
    \end{tabular}
    \vspace{10pt}
    \caption{Post-processing validation results of QA pairs using GPT-4o. Here, \#Total and \#Correctness represent the total number of original QA pairs and the number of correctly answered pairs, respectively.}
    \vspace{10pt}
    \label{tab:post}
\end{table}

\noindent \textbf{Data Splitting} ~ ~
To facilitate a rigorous evaluation for LVLMs, we carefully curate the MMXU-\textit{test} benchmark comprising 1,000 data samples for each of the three categories—``Worsen'', ``Improved'' and ``No Change''. The remaining data forms the MMXU-\textit{dev} dataset. During the selection process, we meticulously ensured that questions derived from the same patient did not appear in both the \textit{dev} and \textit{test} sets, thereby preventing data leakage. Furthermore, we maintained a balanced regional distribution across the training and test sets to preserve the integrity of the evaluation. Finally, the MMXU-\textit{test} contains 3,000 QA pairs, and MMXU-\textit{dev} contains 118,800 QA pairs.

\section{Benchmark Results}
To ensure the professionalism and accuracy of MMXU-\textit{test} benchmark, we recruited a panel of 5 board-certified chest radiology experts to assess it. Following that, we evaluated the performance of several prominent open-source and closed-source LVLMs capable of supporting multi-image VQA on the MMXU-\textit{test} benchmark. Since all the questions from our benchmark are single-choice, we use accuracy as the metric. 

\subsection{Evaluation Models}
For the general domain, we evaluated the following models:

\noindent\textbf{Open-source LVLMs}: Qwen2-VL 2B\& 7B~\cite{bai2023qwen}, DeepSeek-VL 1.3B\& 7B~\cite{lu2024deepseek}, InternVL2 1B\&2B\& 4B\&8B~\cite{chen2024internvl}, IDEFICS2 8B~\cite{laurenccon2024matters} and Llama3.2-Vision 11B~\cite{touvron2023llama}
\noindent\textbf{Closed-source LVLMs}: GPT-4o~\cite{achiam2023gpt} and  Claude-3-5-sonnet

For the medical domain, we evaluated the following models:

\noindent\textbf{Medical LVLMs}: LLaVA-Med-v1.5~\cite{li2024llava}, Med-Flamingo~\cite{moor2023med} and HuatuoGPT-vision~\cite{chen2024huatuogptvisioninjectingmedicalvisual}

\begin{table*}[ht!]
\centering
\small
\begin{tabular}{c|c|c|ccc|cccc}
\toprule
\multirow{2}{*}{\textbf{Source}}&\multirow{2}{*}{\textbf{LVLMs}}& \multirow{2}{*}{\textbf{Size}}& 
\multicolumn{3}{c|}{\textbf{VQA-RAD}} & \multicolumn{4}{c}{\textbf{MMXU-\textit{test}}} \\
\cmidrule{4-10}
& & & \rule{0pt}{8pt}\textbf{Closed} & \textbf{Open} & \textbf{Overall}  
 & \textbf{Worsen} & \textbf{Improved} & \textbf{No change} & \textbf{Overall}  \\
\midrule
\multirow{10}{*}{Open}&\multirow{2}{*}{Qwen2-VL}&
   2B& 0.594&0.380&0.499&\textbf{0.712}&0.258&0.284&0.418\\
& &7B& \textbf{0.745}&0.430&\textbf{0.605}&0.331&0.554&0.500&0.458\\
\cmidrule{2-10}
\rule{0pt}{8pt}&\multirow{2}{*}{DeepSeek-VL}
&1.3B& 0.566&0.250&0.426&0.292&0.294&0.288&0.291\\
& &7B& 0.582&0.300&0.457&0.389&0.310&\underline{0.606}&0.435\\
\cmidrule{2-10}
\rule{0pt}{8pt}&\multirow{4}{*}{InternVL2}
  &1B& 0.490&0.265&0.390& 0.553&0.214&0.113&0.293\\
& &2B& 0.641&0.350&0.512& 0.560&0.182&0.559&0.434\\
& &4B& 0.649&0.370&0.525&0.476&0.571&0.315&0.454\\
& &8B& 0.665&0.480&0.583&0.423&0.483& 0.495 & 0.467 \\
\cmidrule{2-10}
\rule{0pt}{10pt} &IDEFICS2&8B& \underline{0.673}&0.450&0.574& 0.234& 0.570 & \textbf{0.668}&0.491\\
\cmidrule{2-10}
\rule{0pt}{10pt} &Llama3.2-vision&11B& 0.649&\textbf{0.515}&\underline{0.590}&\underline{0.596}&\underline{0.608}&0.356&\underline{0.520}\\
\midrule
\multirow{2}{*}{Closed}& GPT-4o&-&0.578& 0.480&0.534&0.480&\textbf{0.675}&0.559&\textbf{0.571}\\
\cmidrule{2-10}
\rule{0pt}{10pt}& Claude-3.5&-& 0.622&\underline{0.510}&0.572&0.494&0.518&0.493&0.502\\
\bottomrule
\end{tabular}
\caption{Evaluation results of several mainstream open-source and closed-source LVLMs supporting multi-image question answering on the VQA-RAD and MMXU-\textit{test} benchmarks. For open-ended questions in VQA-RAD, we used GPT-4o to evaluate LVLMs' responses. The results in \textbf{bold} and \underline{underlined} represent the best and the second-best
results, respectively.}
\vspace{10pt}
\label{tab:zero-shot}
\end{table*}

\subsection{Human Expert Evaluation}
\label{sec:humanexpertevaluation}
To evaluate the quality of the MMXU-\textit{test} benchmark, we conduct the human expert evaluation with five radiologists. The data from the MMXU-test benchmark was randomly divided into five parts, containing 500, 500, 500, 750, and 750 questions, respectively. We ensured that the three question categories were distributed as evenly as possible within each subset. The evaluation results are presented in Table \ref{tab:expert_evaluation}. Except for Expert 3, all experts achieved an accuracy rate of at least 96.0\%, with an overall accuracy reaching 95.3\%. These findings demonstrate that the MMXU-\textit{test} benchmark is both highly professional and well-structured.


\begin{table}[t]
    \centering\small
    \begin{tabular}{c|ccc}
        \toprule
        \textbf{Expert ID} & \textbf{\#Examples} & \textbf{\#Correctness} & \textbf{Rate} \\
        \midrule
        1 & 500 & 487 & 97.4\% \\
        2 & 500 & 480 & 96.0\% \\
        3 & 500 & 453 & 90.6\% \\
        4 & 750 & 732 & 97.6\% \\
        5 & 750 & 723 & 96.4\% \\
        \midrule
        \textbf{Total} & 3,000 & 2,875 & 95.8\% \\
        \bottomrule
    \end{tabular}
    \caption{Evaluation results of five experienced chest radiology human experts on the MMXU-\textit{test} benchmark.}
    \vspace{10pt}
    \label{tab:expert_evaluation}
\end{table}

\subsection{Model Performance}

We evaluated the performance of several well-known open-source and closed-source models on two benchmarks: VQA-RAD, which focuses on single-image visual question answering, and MMXU-\textit{test}, which emphasizes multi-image difference analysis. The experimental results are presented in Table \ref{tab:zero-shot}. For open-ended questions in VQA-RAD~\cite{Lau2018}, we utilized GPT-4o to compare the model-generated answers with the ground truth, incorporating a certain level of tolerance to assess correctness ~\cite{he2024pefomed}.

Comparing the performance of these models on single-image and multi-image benchmarks, it is clear that \textbf{the accuracy of the same model on overall accuracy on MMXU-\textit{test} is lower than its VQA-RAD closed-ended questions}.
For example, with the Qwen2-VL 7B model, the accuracy on VQA-RAD closed questions reaches an impressive 74.5\%. However, its performance on MMXU-\textit{test} drops significantly to just 45.8\%. This discrepancy highlights the gap between current MedVQA benchmarks and the demands of real-world scenarios, suggesting that models that perform well on public benchmarks may not be effective in supporting clinical diagnosis. Furthermore, as shown in Table \ref{tab:expert_evaluation}, even the best-performing model lags behind human experts by nearly 40\%. This highlights the existing limitations of these models in multi-image MedVQA tasks.
\begin{table}[t]
    \centering\small
    \begin{tabular}{l|ccccc}
        \toprule
        \textbf{Model}  & \textbf{Wors.} & \textbf{Impr.} & \textbf{Noch.} & \textbf{Over.} \\
        \midrule
        Med-Flamingo-7B & 0.289 & 0.304 & 0.251 & 0.281 \\
        Llava-med-7B & 0.293 & 0.334 & 0.323 & 0.317 \\
        HuatuoGPT-Vision-7B & 0.629 & 0.502 & 0.354 & 0.495 \\
        \bottomrule
    \end{tabular}
    \caption{Performance comparison of various medical vision–language models.}
    \vspace{10pt}
    \label{tab:medical_model_comparison}
\end{table}


Additionally, we observe that nearly all open-source models \textbf{demonstrate substantial disparities in accuracy across the three question types} in the MMXU-\textit{test}, indicating inherent biases in how these models interpret disease progression. This phenomenon is especially pronounced in smaller models---for instance, Qwen2-VL 2B achieves an accuracy of 71.2\% on ``Worsen'' questions, yet performs below 30\% on the remaining two types. Evidently, the model exhibits a strong inclination toward outcomes associated with disease deterioration. In contrast, such biases are less prominent in closed-source models. \textbf{With increasing model size, overall accuracy improves}, and the extent of progression-related bias tends to decrease. Despite being general-purpose in nature, these models still produce divergent results. Notably, smaller models tend to yield relatively consistent performance, whereas discrepancies become more pronounced once the model size surpasses 7B. For example, models such as DeepSeek-VL 1.3B and InternVL2 1B, or Qwen2-VL 2B and InternVL2 2B, exhibit similar capabilities. Conversely, larger models like DeepSeek-VL 7B, InternVL2 8B, and IDEFICS2 8B demonstrate marked differences in their performance.

As shown in Table \ref{tab:medical_model_comparison}, we also evaluated three medical models. For models restricted to single-image inputs, we concatenated two images to facilitate evaluation. The results, summarized in the table below, indicate that HuatuoGPT-Vision outperformed other models of similar size and achieved performance comparable to the commercial model Claude 3.5.

\section{MedRecord-Augmented Generation}

Through prior experiments, we pinpointed that current LVLMs face significant challenges in identifying disease progression. To address this challenge, we propose a novel approach, MedRecord-Augmented Generation (MAG). In routine diagnostic practice, physicians often rely on a patient’s historical records to inform their analysis and diagnosis of current conditions. To replicate this process, we integrate historical records directly into the prompt as contextual information. Our study investigates the effectiveness of global reports derived from historical images, alongside regional reports related to the specified questions, which we categorize as global and regional historical records. 

\begin{table}[t]
\centering
\small
\resizebox{0.5\textwidth}{!}{%
\begin{tabular}{lcccc}
\toprule
\textbf{Method}& \textbf{Worsen} & \textbf{Improved} & \textbf{No change} & \textbf{Overall} \\
\midrule
\rowcolor[HTML]{EDEDED}\multicolumn{5}{c}{\textit{\textbf{InternVL2 8B}}}\\
- &0.423&0.483&0.495&0.467\\
+ G-MRec&0.440&0.644&0.838&0.641\\
+ R-MRec&\textbf{0.483}&\textbf{0.830}&\textbf{0.842}&\textbf{0.718}\\
\midrule
\rowcolor[HTML]{EDEDED}\multicolumn{5}{c}{\textit{\textbf{Qwen2-VL 7B}}}\\
- &0.331&0.544&0.500&0.458\\
+ G-MRec&0.427&0.683&0.784&0.631\\
+ R-MRec&\textbf{0.458}&\textbf{0.755}&\textbf{0.805}&\textbf{0.673}\\
\midrule
\rowcolor[HTML]{EDEDED}\multicolumn{5}{c}{\textit{\textbf{GPT-4o}}}\\
- &\textbf{0.480}&0.675&0.559&0.571\\
+ G-MRec&0.380&0.760&0.629&0.590\\
+ R-MRec&0.374&\textbf{0.802}&\textbf{0.765}&\textbf{0.647}\\
\midrule
\rowcolor[HTML]{EDEDED}\multicolumn{5}{c}{\textit{\textbf{HuatuoGPT-Vision 7B}}}\\
- &0.629&0.502&0.354&0.495\\
+ G-MRec&0.468&0.660&0.792&0.640\\
+ R-MRec&\textbf{0.490}&\textbf{0.712}&\textbf{0.775}&\textbf{0.659}\\
\midrule
\rowcolor[HTML]{EDEDED}\multicolumn{5}{c}{\textit{\textbf{Llava-med-v1.5-mistral 7B}}}\\
- &0.293&0.334&0.323&0.317\\
+ G-MRec&0.378&0.468&\textbf{0.708}&\textbf{0.518}\\
+ R-MRec&\textbf{0.396}&\textbf{0.493}&0.647&0.512\\
\bottomrule
\end{tabular}
}
\caption{Results of MedRecord-augmented generation on the MMXU-\textit{test} benchmark without fine-tuning. G-MRec and R-MRec denote generation augmentation using global and regional historical records, respectively.}
\vspace{10pt}
\label{tab:zero-shot-mag}
\end{table}

\subsection{Effectiveness of MAG}

To more comprehensively evaluate the effectiveness of MAG, we conducted tests across general-domain open-source models, commercial models, and medical models. The outcomes of these experiments are summarized in Table \ref{tab:zero-shot-mag}. It is evident that providing historical records significantly enhances overall model accuracy, regardless of the model type. With access to historical records, open-source models achieved performance comparable to that of closed-source commercial models.

With the exception of GPT-4o and Llava-med-v1.5, incorporating medical records led to improvements across all three question types, suggesting that these models are capable of understanding and reasoning from historical data rather than merely repeating it. Furthermore, it is noticeable that regional historical records yield greater improvements than global records in nearly all models. This may be due to the models’ limited ability to analyze contextual and region-specific nuances, which makes it challenging to pinpoint the most relevant information. By supplying precise, localized historical data, the models are better equipped to extract meaningful insights.

The extent to which historical records contribute to performance improvement varies across different question categories. As illustrated in Table \ref{tab:zero-shot-mag}, historical records notably enhance performances in the ``Improved'' and ``No Change'' categories, while providing only minimal benefit in the ``Worsen'' category. In fact, for GPT-4o, performance in this category even deteriorated. This indicates that the use of historical records may \textbf{not effectively address biases} in a high-quality and robust manner.

\begin{table}[t]
\centering
\small
\resizebox{0.48\textwidth}{!}{%
\begin{tabular}{lcccc}
\toprule
\textbf{Method }& \textbf{Worsen} & \textbf{Improved} & \textbf{No change} & \textbf{Overall} \\
\midrule
\rowcolor[HTML]{EDEDED}\multicolumn{5}{c}{\textit{\textbf{InternVL2 8B + 20\% MMXU-\textit{dev}}}}\\
-&0.834&0.802&0.810&0.815\\
+ G-MRec&0.856&0.841&0.843&0.847\\
+ R-MRec&0.866&0.830&0.848&0.848\\
\midrule
\rowcolor[HTML]{EDEDED}\multicolumn{5}{c}{\textit{\textbf{InternVL2 8B + 40\% MMXU-\textit{dev}}}}\\
-&0.838&0.813&0.839&0.830\\
+ G-MRec&0.867&0.857&0.867&0.864\\
+ R-MRec&0.868&0.845&0.874&0.862\\
\midrule
\rowcolor[HTML]{EDEDED}\multicolumn{5}{c}{\textit{\textbf{InternVL2 8B + 60\% MMXU-\textit{dev}}}}\\
-&0.855&0.822&0.841&0.839\\
+ G-MRec&0.884&0.855&0.873&0.871\\
+ R-MRec&0.880&0.855&0.869&0.868\\
\midrule
\rowcolor[HTML]{EDEDED}\multicolumn{5}{c}{\textit{\textbf{InternVL2 8B + 80\% MMXU-\textit{dev}}}}\\
-&0.857&0.814&0.866&0.846\\
+ G-MRec&0.881&0.861&0.884&0.876\\
+ R-MRec&0.873&0.869&0.881&0.875\\
\midrule
\rowcolor[HTML]{EDEDED}\multicolumn{5}{c}{\textit{\textbf{InternVL2 8B + 100\% MMXU-\textit{dev}}}}\\
-&0.860&0.840&0.852&0.851\\
+ G-MRec&0.887&0.878&0.882&0.883\\
+ R-MRec&0.884&0.871&0.888&0.881\\
\bottomrule
\end{tabular}
}
\caption{Fine-tuning Results of InternVL2 8B on MMXU-\textit{dev} dataset with MAG Method. G-MRec and R-MRec denote generation augmentation using global and regional historical records, respectively.}
\vspace{10pt}
\label{fig:finetuned}
\end{table}

\subsection{MAG Fine-tuning on MMXU-\textit{dev}}

Furthermore, we assess the efficacy of MMXU-\textit{dev} dataset and our proposed MAG method by fine-tuning the InternVL2 8B. The model was fine-tuned using 20\%, 40\%, 60\%, 80\%, and 100\% of the MMXU-\textit{dev} dataset. We evaluate the performance on MMXU-\textit{test} benchmark. The corresponding fine-tuning results are presented in Table \ref{fig:finetuned}.

We observed that even with just 20\% of the MMXU-\textit{dev}, the model achieved substantial improvements in accuracy across all three problem categories, and \textbf{the bias was almost eliminated}. As the dataset size expanded, the model's overall performance continued to improve. When utilizing the full 100\% of MMXU-\textit{dev}, the performance gap between the model and human experts narrowed to approximately 10\%. This clearly underscores the efficacy of the MMXU-\textit{dev} dataset we developed. Additionally, we found that the MAG method continues to deliver significant improvements after model fine-tuning. 
Notably, as the volume of training data increases,\textbf{ the enhancement effect of global historical records surpasses that of regional historical records}.
Fine-tuning with larger datasets enables the model to more effectively capture relevant information, resulting in more comprehensive and refined final outputs.

\subsection{Ablation Study on MAG}

To assess whether the MAG model disproportionately relies on report content, we evaluate the performance of InternVL2 8B using only global and regional historical records—excluding imaging data—for question answering. As shown in Table~\ref{tab:ablation_mrec}, the experimental results indicate a marked performance decline in the zero-shot setting when relying solely on textual input, as compared to utilizing both textual and visual modalities.

This degradation is expected and underscores the critical role of imaging data. Several factors contribute to this dependency: clinical reports often lack detailed descriptions of prior findings or may contain inconsistencies. In cases involving disease progression or resolution, textual narratives alone frequently fall short of conveying the nuance required to assess subtle changes. Image comparison is indispensable for accurately gauging such progression. For example, if a prior report notes an anomaly still visible in the current scan, the text alone cannot determine whether the condition has improved, deteriorated, or remained stable. Furthermore, if a subtle lesion was previously overlooked and thus omitted from the report, the model might incorrectly classify it as a novel finding, despite its earlier presence in the images.

\begin{table}[t]
    \centering\small
    \resizebox{\linewidth}{!}{%
    \begin{tabular}{l|cccc}
        \toprule
        \textbf{Method} & \textbf{Worsen} & \textbf{Improved} & \textbf{No Change} & \textbf{Overall} \\
        \midrule
        InternVL2 8B & 0.423 & 0.483 & 0.495 & 0.467 \\
        $+\,$G-MRec & 0.443 & 0.644 & 0.838 & 0.641 \\
        $+\,$G-MRec w/o PI & 0.415 & 0.510 & 0.835 & 0.573 \\
        $+\,$R-MRec & 0.483 & 0.830 & 0.842 & 0.718 \\
        $+\,$R-MRec w/o PI& 0.364 & 0.593 & 0.823 & 0.593 \\
        \bottomrule
    \end{tabular}%
    }
    \caption{Ablation study on the impact of prior-image information and historical records. Here, w/o PI denotes the removal of the prior image when using the MAG method.}
    \vspace{10pt}
    \label{tab:ablation_mrec}
\end{table}

\begin{figure*}[t]
  \centering
  \includegraphics[width=2.05\columnwidth]{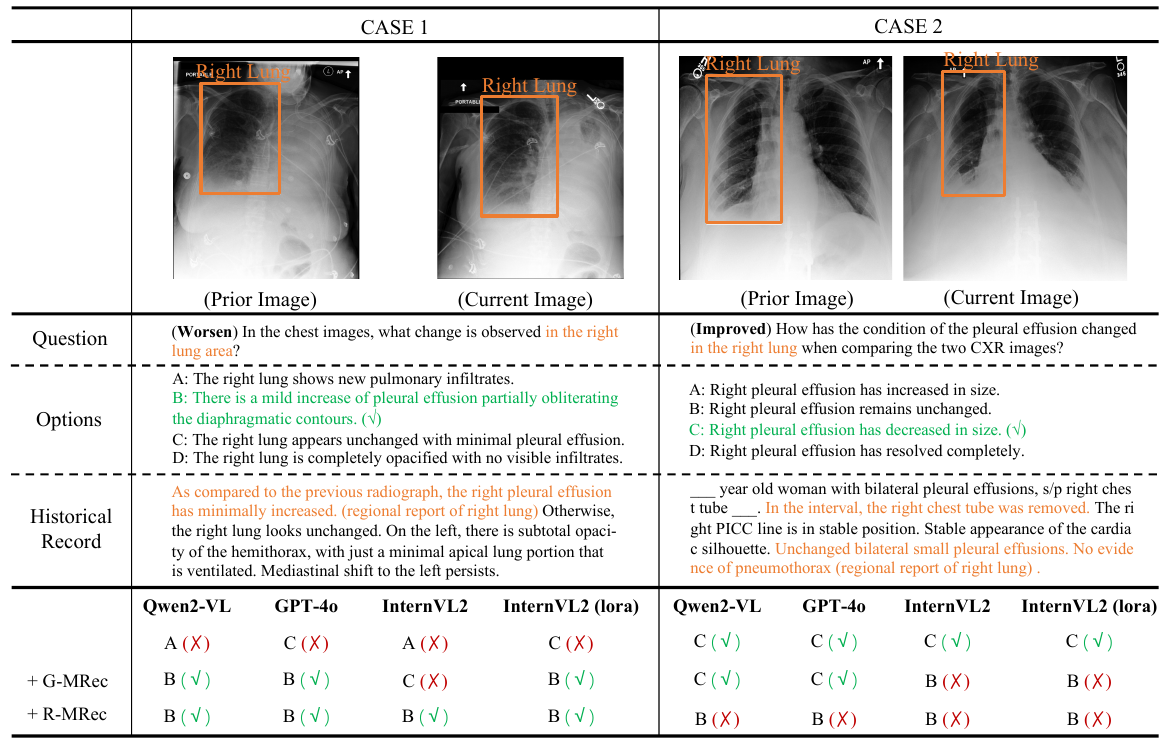}
  \caption{Two Examples from MMXU answered by some LVLMs. The left image shows successful cases with medical records enhancing the answers. The right image shows failed cases where historical records mislead the answers.}
  \vspace{10pt}
  \label{fig:case}
\end{figure*}

\subsection{Case Study}

In Figure \ref{fig:case}, we present two answer samples provided by GPT-4o, InternVL2 8B, Qwen2-VL 7B, and fine-tuned InternVL2 8B, evaluated under various historical record strategies. Case 1 illustrates a successful scenario where all models adjusted their responses accurately when regional historical records were incorporated.  However, the zero-shot InternVL model misinterpreted critical information, primarily due to the global records containing terms such as ``unchanged'' and ``minimally,'' which led to an erroneous conclusion. Fine-tuning effectively mitigated this issue. In Case 2, historical records resulted in misleading predictions. When historical information was absent, all models delivered correct answers. However, the current right pleural effusion had diminished in size, and prior diagnoses indicated no changes. The inclusion of historical records resulted in an incorrect response, and fine-tuning did not yield an improvement in this case. For more case studies, please refer to Appendix \ref{app:more-case-study}.

\section{Conclusion}

In this paper, we introduce MMXU, a dataset designed for multi-modal and multi-X-ray understanding in MedVQA. First, we propose a benchmark, MMXU-\textit{test}, and invite five chest X-ray experts to evaluate the performance. Then, we conduct evaluations using several well-known open-source and closed-source large vision-language models (LVLMs) that support multi-image VQA. The experimental results indicate that even the best-performing models exhibit a significant performance gap—nearly 40\%—compared to human experts. To bridge this gap, we propose the MAG method, which leverages historical records to enhance the understanding of disease progression. We further evaluate its performance both without fine-tuning and with fine-tuning on MMXU-\textit{dev} and the experiment confirm the effectiveness.

\section*{Limitation}
Although we have carefully designed our MMXU-\textit{test} benchmark and established the MMXU-\textit{dev} dataset, there are still some limitations: 1) Our dataset is based on MIMIC-CXR, a chest X-ray dataset, which somewhat limits its generalization when applied to other datasets. 2) We have proposed the MAG method, inspired by clinical scenarios, to validate the effectiveness of historical historical records in enhancing LVLMs' medical responses. Although we aim to replicate clinical scenarios as closely as possible, the scarcity of data means we can only use previous reports as historical records for research purposes.

\section*{Acknowledgment}
This work was supported by the National Natural Science Foundation of China (No. 62301311) and the Fundamental Research Funds for the Central Universities (No. YG2022QN029).

\newpage
\bibliography{custom}

\begin{thebibliography}{41}
\providecommand{\natexlab}[1]{#1}

\bibitem[{Achiam et~al.(2023)Achiam, Adler, Agarwal, Ahmad, Akkaya, Aleman, Almeida, Altenschmidt, Altman, Anadkat et~al.}]{achiam2023gpt}
Josh Achiam, Steven Adler, Sandhini Agarwal, Lama Ahmad, Ilge Akkaya, Florencia~Leoni Aleman, Diogo Almeida, Janko Altenschmidt, Sam Altman, Shyamal Anadkat, et~al. 2023.
\newblock Gpt-4 technical report.
\newblock \emph{arXiv preprint arXiv:2303.08774}.

\bibitem[{Alayrac et~al.(2022)Alayrac, Donahue, Luc, Miech, Barr, Hasson, Lenc, Mensch, Millican, Reynolds et~al.}]{alayrac2022flamingo}
Jean-Baptiste Alayrac, Jeff Donahue, Pauline Luc, Antoine Miech, Iain Barr, Yana Hasson, Karel Lenc, Arthur Mensch, Katherine Millican, Malcolm Reynolds, et~al. 2022.
\newblock Flamingo: a visual language model for few-shot learning.
\newblock \emph{Advances in neural information processing systems}, 35:23716--23736.

\bibitem[{Bae et~al.(2024)Bae, Kyung, Ryu, Cho, Lee, Kweon, Oh, Ji, Chang, Kim et~al.}]{bae2024ehrxqa}
Seongsu Bae, Daeun Kyung, Jaehee Ryu, Eunbyeol Cho, Gyubok Lee, Sunjun Kweon, Jungwoo Oh, Lei Ji, Eric Chang, Tackeun Kim, et~al. 2024.
\newblock Ehrxqa: A multi-modal question answering dataset for electronic health records with chest x-ray images.
\newblock \emph{Advances in Neural Information Processing Systems}, 36.

\bibitem[{Bai et~al.(2023)Bai, Bai, Yang, Wang, Tan, Wang, Lin, Zhou, and Zhou}]{bai2023qwen}
Jinze Bai, Shuai Bai, Shusheng Yang, Shijie Wang, Sinan Tan, Peng Wang, Junyang Lin, Chang Zhou, and Jingren Zhou. 2023.
\newblock Qwen-vl: A versatile vision-language model for understanding, localization, text reading, and beyond.
\newblock \emph{arXiv preprint arXiv:2308.12966}, 1(2):3.

\bibitem[{Bu et~al.(2024)Bu, Song, Li, and Dai}]{bu-etal-2024-dynamic}
Shenshen Bu, Yujie Song, Taiji Li, and Zhiming Dai. 2024.
\newblock Dynamic knowledge prompt for chest {X}-ray report generation.
\newblock In \emph{LREC:2024:main}, pages 5425--5436, Torino, Italia. ELRA and ICCL.

\bibitem[{Chen et~al.(2024{\natexlab{a}})Chen, Cai, Ji, Wang, Liu, Wang, Hou, and Wang}]{chen2024huatuogpto1medicalcomplexreasoning}
Junying Chen, Zhenyang Cai, Ke~Ji, Xidong Wang, Wanlong Liu, Rongsheng Wang, Jianye Hou, and Benyou Wang. 2024{\natexlab{a}}.
\newblock Huatuogpt-o1, towards medical complex reasoning with llms.
\newblock \emph{arXiv preprint arXiv:2412.18925}.

\bibitem[{Chen et~al.(2024{\natexlab{b}})Chen, Ouyang, Gao, Chen, Chen, Wang, Zhang, Cai, Ji, Yu, Wan, and Wang}]{chen2024huatuogptvisioninjectingmedicalvisual}
Junying Chen, Ruyi Ouyang, Anningzhe Gao, Shunian Chen, Guiming~Hardy Chen, Xidong Wang, Ruifei Zhang, Zhenyang Cai, Ke~Ji, Guangjun Yu, Xiang Wan, and Benyou Wang. 2024{\natexlab{b}}.
\newblock \href {https://arxiv.org/abs/2406.19280} {Huatuogpt-vision, towards injecting medical visual knowledge into multimodal llms at scale}.
\newblock \emph{Preprint}, arXiv:2406.19280.

\bibitem[{Chen et~al.(2024{\natexlab{c}})Chen, Wu, Wang, Su, Chen, Xing, Zhong, Zhang, Zhu, Lu et~al.}]{chen2024internvl}
Zhe Chen, Jiannan Wu, Wenhai Wang, Weijie Su, Guo Chen, Sen Xing, Muyan Zhong, Qinglong Zhang, Xizhou Zhu, Lewei Lu, et~al. 2024{\natexlab{c}}.
\newblock Internvl: Scaling up vision foundation models and aligning for generic visual-linguistic tasks.
\newblock In \emph{Proceedings of the IEEE/CVF Conference on Computer Vision and Pattern Recognition}, pages 24185--24198.

\bibitem[{Dai et~al.(2023)Dai, Li, Li, Tiong, Zhao, Wang, Li, Fung, and Hoi}]{dai2023instructblip}
Wenliang Dai, Junnan Li, D~Li, AMH Tiong, J~Zhao, W~Wang, B~Li, P~Fung, and S~Hoi. 2023.
\newblock Instructblip: Towards general-purpose vision-language models with instruction tuning. arxiv 2023.
\newblock \emph{arXiv preprint arXiv:2305.06500}, 2.

\bibitem[{He et~al.(2024)He, Li, Liu, Zhao, and Zhong}]{he2024pefomed}
Jinlong He, Pengfei Li, Gang Liu, Zixu Zhao, and Shenjun Zhong. 2024.
\newblock Pefomed: Parameter efficient fine-tuning on multimodal large language models for medical visual question answering.
\newblock \emph{arXiv preprint arXiv:2401.02797}.

\bibitem[{Hong et~al.(2024)Hong, Wang, Ding, Yu, Lv, Wang, Cheng, Huang, Ji, Xue et~al.}]{hong2024cogvlm2}
Wenyi Hong, Weihan Wang, Ming Ding, Wenmeng Yu, Qingsong Lv, Yan Wang, Yean Cheng, Shiyu Huang, Junhui Ji, Zhao Xue, et~al. 2024.
\newblock Cogvlm2: Visual language models for image and video understanding.
\newblock \emph{arXiv preprint arXiv:2408.16500}.

\bibitem[{Hu et~al.(2023)Hu, Gu, An, Zhang, Liu, Kobayashi, Harada, Summers, and Zhu}]{hu2023medical}
Xinyue Hu, L~Gu, Q~An, M~Zhang, L~Liu, K~Kobayashi, T~Harada, R~Summers, and Y~Zhu. 2023.
\newblock Medical-diff-vqa: a large-scale medical dataset for difference visual question answering on chest x-ray images.
\newblock \emph{PhysioNet}, 12:13.

\bibitem[{Hu et~al.(2024)Hu, Li, Lu, Shao, He, Qiao, and Luo}]{hu2024omnimedvqa}
Yutao Hu, Tianbin Li, Quanfeng Lu, Wenqi Shao, Junjun He, Yu~Qiao, and Ping Luo. 2024.
\newblock Omnimedvqa: A new large-scale comprehensive evaluation benchmark for medical lvlm.
\newblock In \emph{Proceedings of the IEEE/CVF Conference on Computer Vision and Pattern Recognition}, pages 22170--22183.

\bibitem[{Huang et~al.(2025)Huang, Han, L, Li, Wu, and Zhang}]{huang-etal-2025-cmeaa}
Xiyang Huang, Yingjie Han, Yx~L, Runzhi Li, Pengcheng Wu, and Kunli Zhang. 2025.
\newblock \href {2025.coling-main.571/} {{C}m{EAA}: Cross-modal enhancement and alignment adapter for radiology report generation}.
\newblock In \emph{COLING:2025:main}, pages 8546--8556, Abu Dhabi, UAE. acl.

\bibitem[{Johnson et~al.(2019)Johnson, Pollard, Berkowitz, Greenbaum, Lungren, Deng, Mark, and Horng}]{johnson2019mimic}
Alistair~EW Johnson, Tom~J Pollard, Seth~J Berkowitz, Nathaniel~R Greenbaum, Matthew~P Lungren, Chih-ying Deng, Roger~G Mark, and Steven Horng. 2019.
\newblock Mimic-cxr, a de-identified publicly available database of chest radiographs with free-text reports.
\newblock \emph{Scientific data}, 6(1):317.

\bibitem[{Lau et~al.(2018)Lau, Gayen, Ben, and Demner-Fushman}]{Lau2018}
Jason Lau, Soumya Gayen, Asma Ben, and Dina Demner-Fushman. 2018.
\newblock \href {https://doi.org/10.1038/sdata.2018.251} {A dataset of clinically generated visual questions and answers about radiology images}.
\newblock \emph{Scientific Data}, 5(1):180251.

\bibitem[{Lauren{\c{c}}on et~al.(2024)Lauren{\c{c}}on, Tronchon, Cord, and Sanh}]{laurenccon2024matters}
Hugo Lauren{\c{c}}on, L{\'e}o Tronchon, Matthieu Cord, and Victor Sanh. 2024.
\newblock What matters when building vision-language models?
\newblock \emph{arXiv preprint arXiv:2405.02246}.

\bibitem[{Li et~al.(2024{\natexlab{a}})Li, Wong, Zhang, Usuyama, Liu, Yang, Naumann, Poon, and Gao}]{li2024llava}
Chunyuan Li, Cliff Wong, Sheng Zhang, Naoto Usuyama, Haotian Liu, Jianwei Yang, Tristan Naumann, Hoifung Poon, and Jianfeng Gao. 2024{\natexlab{a}}.
\newblock Llava-med: Training a large language-and-vision assistant for biomedicine in one day.
\newblock \emph{Advances in Neural Information Processing Systems}, 36.

\bibitem[{Li et~al.(2023{\natexlab{a}})Li, Li, Savarese, and Hoi}]{li2023blip}
Junnan Li, Dongxu Li, Silvio Savarese, and Steven Hoi. 2023{\natexlab{a}}.
\newblock Blip-2: Bootstrapping language-image pre-training with frozen image encoders and large language models.
\newblock In \emph{International conference on machine learning}, pages 19730--19742. PMLR.

\bibitem[{Li et~al.(2024{\natexlab{b}})Li, Zhang, Wang, Zhong, Chen, Chu, Liu, and Jia}]{li2024mini}
Yanwei Li, Yuechen Zhang, Chengyao Wang, Zhisheng Zhong, Yixin Chen, Ruihang Chu, Shaoteng Liu, and Jiaya Jia. 2024{\natexlab{b}}.
\newblock Mini-gemini: Mining the potential of multi-modality vision language models.
\newblock \emph{arXiv preprint arXiv:2403.18814}.

\bibitem[{Li et~al.(2023{\natexlab{b}})Li, Li, Zhang, Dan, Jiang, and Zhang}]{li2023chatdoctor}
Yunxiang Li, Zihan Li, Kai Zhang, Ruilong Dan, Steve Jiang, and You Zhang. 2023{\natexlab{b}}.
\newblock Chatdoctor: A medical chat model fine-tuned on a large language model meta-ai (llama) using medical domain knowledge.
\newblock \emph{Cureus}, 15(6).

\bibitem[{Liu et~al.(2021)Liu, Zhan, Xu, Ma, Yang, and Wu}]{liu2021slake}
Bo~Liu, Li-Ming Zhan, Li~Xu, Lin Ma, Yan Yang, and Xiao-Ming Wu. 2021.
\newblock Slake: A semantically-labeled knowledge-enhanced dataset for medical visual question answering.
\newblock In \emph{2021 IEEE 18th International Symposium on Biomedical Imaging (ISBI)}, pages 1650--1654. IEEE.

\bibitem[{Liu et~al.(2024)Liu, Zou, Zhan, Lu, Dong, Chen, Xie, Cao, Wu, and Fu}]{liu2024gemex}
Bo~Liu, Ke~Zou, Liming Zhan, Zexin Lu, Xiaoyu Dong, Yidi Chen, Chengqiang Xie, Jiannong Cao, Xiao-Ming Wu, and Huazhu Fu. 2024.
\newblock Gemex: A large-scale, groundable, and explainable medical vqa benchmark for chest x-ray diagnosis.
\newblock \emph{arXiv preprint arXiv:2411.16778}.

\bibitem[{Liu et~al.(2023)Liu, Li, Wu, and Lee}]{liu2023llava}
Haotian Liu, Chunyuan Li, Qingyang Wu, and Yong~Jae Lee. 2023.
\newblock Visual instruction tuning.

\bibitem[{Lorkowski and Pokorski(2022)}]{lorkowski2022medical}
Jacek Lorkowski and Mieczyslaw Pokorski. 2022.
\newblock Medical records: A historical narrative.
\newblock \emph{Biomedicines}, 10(10):2594.

\bibitem[{Lu et~al.(2024)Lu, Liu, Zhang, Wang, Dong, Liu, Sun, Ren, Li, Yang et~al.}]{lu2024deepseek}
Haoyu Lu, Wen Liu, Bo~Zhang, Bingxuan Wang, Kai Dong, Bo~Liu, Jingxiang Sun, Tongzheng Ren, Zhuoshu Li, Hao Yang, et~al. 2024.
\newblock Deepseek-vl: towards real-world vision-language understanding.
\newblock \emph{arXiv preprint arXiv:2403.05525}.

\bibitem[{Moor et~al.(2023)Moor, Huang, Wu, Yasunaga, Dalmia, Leskovec, Zakka, Reis, and Rajpurkar}]{moor2023med}
Michael Moor, Qian Huang, Shirley Wu, Michihiro Yasunaga, Yash Dalmia, Jure Leskovec, Cyril Zakka, Eduardo~Pontes Reis, and Pranav Rajpurkar. 2023.
\newblock Med-flamingo: a multimodal medical few-shot learner.
\newblock In \emph{Machine Learning for Health (ML4H)}, pages 353--367. PMLR.

\bibitem[{Nisar et~al.(2025)Nisar, Anwar, Jiang, Parida, Sanchez-Jacob, Nath, Roth, and Linguraru}]{10.1007/978-3-031-73471-7_10}
Hareem Nisar, Syed~Muhammad Anwar, Zhifan Jiang, Abhijeet Parida, Ramon Sanchez-Jacob, Vishwesh Nath, Holger~R. Roth, and Marius~George Linguraru. 2025.
\newblock D-rax: Domain-specific radiologic assistant leveraging multi-modal data and expert model predictions.
\newblock In \emph{Foundation Models for General Medical AI}, pages 91--102, Cham. Springer Nature Switzerland.

\bibitem[{Rajbhandari et~al.(2020)Rajbhandari, Rasley, Ruwase, and He}]{rajbhandari2020zero}
Samyam Rajbhandari, Jeff Rasley, Olatunji Ruwase, and Yuxiong He. 2020.
\newblock Zero: Memory optimizations toward training trillion parameter models.
\newblock In \emph{SC20: International Conference for High Performance Computing, Networking, Storage and Analysis}, pages 1--16. IEEE.

\bibitem[{Saeed(2024)}]{saeed-2024-medifact}
Nadia Saeed. 2024.
\newblock \href {https://doi.org/10.18653/v1/2024.clinicalnlp-1.31} {{M}edi{F}act at {MEDIQA}-{M}3{G} 2024: Medical question answering in dermatology with multimodal learning}.
\newblock In \emph{CLINICALNLP:2024:1}, pages 339--345, Mexico City, Mexico. acl.

\bibitem[{Sun et~al.(2024)Sun, Qin, Fu, Wang, and Tao}]{sun-etal-2024-self}
Guohao Sun, Can Qin, Huazhu Fu, Linwei Wang, and Zhiqiang Tao. 2024.
\newblock \href {https://doi.org/10.18653/v1/2024.emnlp-main.1119} {Self-training large language and vision assistant for medical question answering}.
\newblock In \emph{EMNLP:2024:main}, pages 20052--20060, Miami, Florida, USA. acl.

\bibitem[{Touvron et~al.(2023)Touvron, Lavril, Izacard, Martinet, Lachaux, Lacroix, Rozi{\`e}re, Goyal, Hambro, Azhar et~al.}]{touvron2023llama}
Hugo Touvron, Thibaut Lavril, Gautier Izacard, Xavier Martinet, Marie-Anne Lachaux, Timoth{\'e}e Lacroix, Baptiste Rozi{\`e}re, Naman Goyal, Eric Hambro, Faisal Azhar, et~al. 2023.
\newblock Llama: Open and efficient foundation language models.
\newblock \emph{arXiv preprint arXiv:2302.13971}.

\bibitem[{Wang et~al.(2023)Wang, Lv, Yu, Hong, Qi, Wang, Ji, Yang, Zhao, Song et~al.}]{wang2023cogvlm}
Weihan Wang, Qingsong Lv, Wenmeng Yu, Wenyi Hong, Ji~Qi, Yan Wang, Junhui Ji, Zhuoyi Yang, Lei Zhao, Xixuan Song, et~al. 2023.
\newblock Cogvlm: Visual expert for pretrained language models.
\newblock \emph{arXiv preprint arXiv:2311.03079}.

\bibitem[{Wu et~al.(2024{\natexlab{a}})Wu, Lin, Zhang, Zhang, Xie, and Wang}]{wu2024pmc}
Chaoyi Wu, Weixiong Lin, Xiaoman Zhang, Ya~Zhang, Weidi Xie, and Yanfeng Wang. 2024{\natexlab{a}}.
\newblock Pmc-llama: toward building open-source language models for medicine.
\newblock \emph{Journal of the American Medical Informatics Association}, page ocae045.

\bibitem[{Wu et~al.(2021)Wu, Agu, Lourentzou, Sharma, Paguio, Yao, Dee, Mitchell, Kashyap, Giovannini et~al.}]{wu2021chest}
Joy~T Wu, Nkechinyere~N Agu, Ismini Lourentzou, Arjun Sharma, Joseph~A Paguio, Jasper~S Yao, Edward~C Dee, William Mitchell, Satyananda Kashyap, Andrea Giovannini, et~al. 2021.
\newblock Chest imagenome dataset for clinical reasoning.
\newblock \emph{arXiv preprint arXiv:2108.00316}.

\bibitem[{Wu et~al.(2024{\natexlab{b}})Wu, Guo, Luo, Wang, Wang, Yang, Zhu, and Ding}]{wu2024medical}
Zheng Wu, Kehua Guo, Entao Luo, Tian Wang, Shoujin Wang, Yi~Yang, Xiangyuan Zhu, and Rui Ding. 2024{\natexlab{b}}.
\newblock Medical long-tailed learning for imbalanced data: bibliometric analysis.
\newblock \emph{Computer Methods and Programs in Biomedicine}, page 108106.

\bibitem[{Ye et~al.(2024)Ye, Xu, Ye, Yan, Hu, Liu, Qian, Zhang, and Huang}]{ye2024mplug}
Qinghao Ye, Haiyang Xu, Jiabo Ye, Ming Yan, Anwen Hu, Haowei Liu, Qi~Qian, Ji~Zhang, and Fei Huang. 2024.
\newblock mplug-owl2: Revolutionizing multi-modal large language model with modality collaboration.
\newblock In \emph{Proceedings of the IEEE/CVF Conference on Computer Vision and Pattern Recognition}, pages 13040--13051.

\bibitem[{Ye et~al.(2020)Ye, Yao, Xue, and Li}]{ye2020weakly}
Wenwu Ye, Jin Yao, Hui Xue, and Yi~Li. 2020.
\newblock \href {https://arxiv.org/abs/2005.14480} {Weakly supervised lesion localization with probabilistic-cam pooling}.
\newblock \emph{Preprint}, arXiv:2005.14480.

\bibitem[{Yin et~al.(2025)Yin, Zhou, Wang, Wu, and Hao}]{yin-etal-2025-kia}
Heng Yin, Shanlin Zhou, Pandong Wang, Zirui Wu, and Yongtao Hao. 2025.
\newblock \href {2025.coling-main.276/} {{KIA}: Knowledge-guided implicit vision-language alignment for chest {X}-ray report generation}.
\newblock In \emph{COLING:2025:main}, pages 4096--4108, Abu Dhabi, UAE. acl.

\bibitem[{Zhou and Wang(2024)}]{zhou-wang-2024-divide}
Yuanpin Zhou and Huogen Wang. 2024.
\newblock \href {https://doi.org/10.18653/v1/2024.emnlp-main.433} {Divide and conquer radiology report generation via observation level fine-grained pretraining and prompt tuning}.
\newblock In \emph{EMNLP:2024:main}, pages 7597--7610, Miami, Florida, USA. acl.

\bibitem[{Zhu et~al.(2023)Zhu, Chen, Shen, Li, and Elhoseiny}]{zhu2023minigpt}
Deyao Zhu, Jun Chen, Xiaoqian Shen, Xiang Li, and Mohamed Elhoseiny. 2023.
\newblock Minigpt-4: Enhancing vision-language understanding with advanced large language models.
\newblock \emph{arXiv preprint arXiv:2304.10592}.

\end{thebibliography}

\appendix

\begin{figure*}[htbp]
    \centering
    \subfigure[]{
        \includegraphics[width=0.45\textwidth,height=5cm]{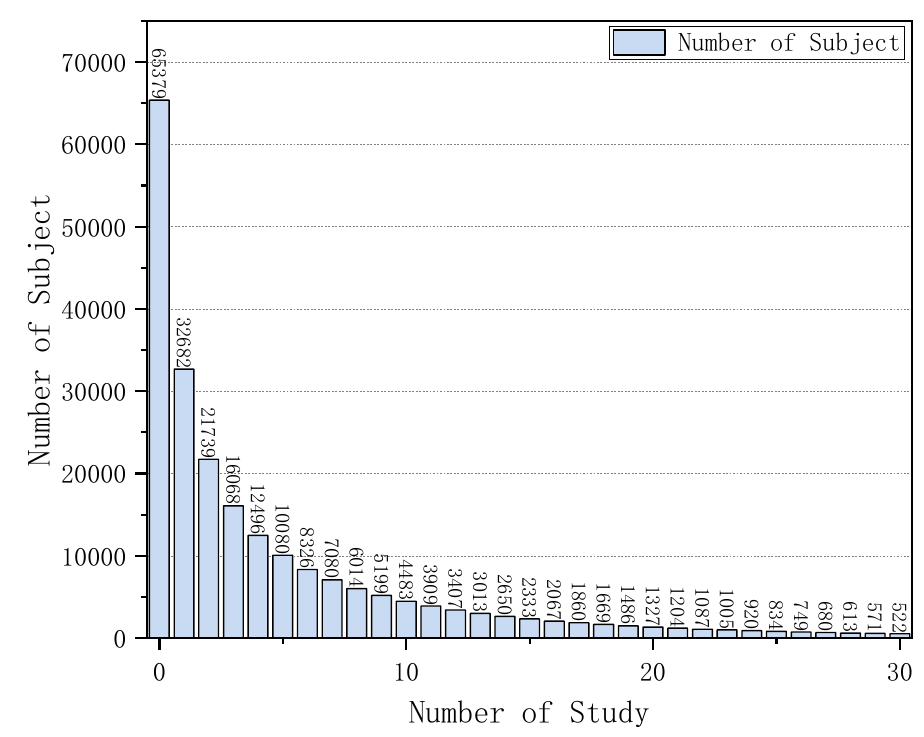}
        \label{fig:m-3}
    }
    \subfigure[]{
        \includegraphics[width=0.45\textwidth]{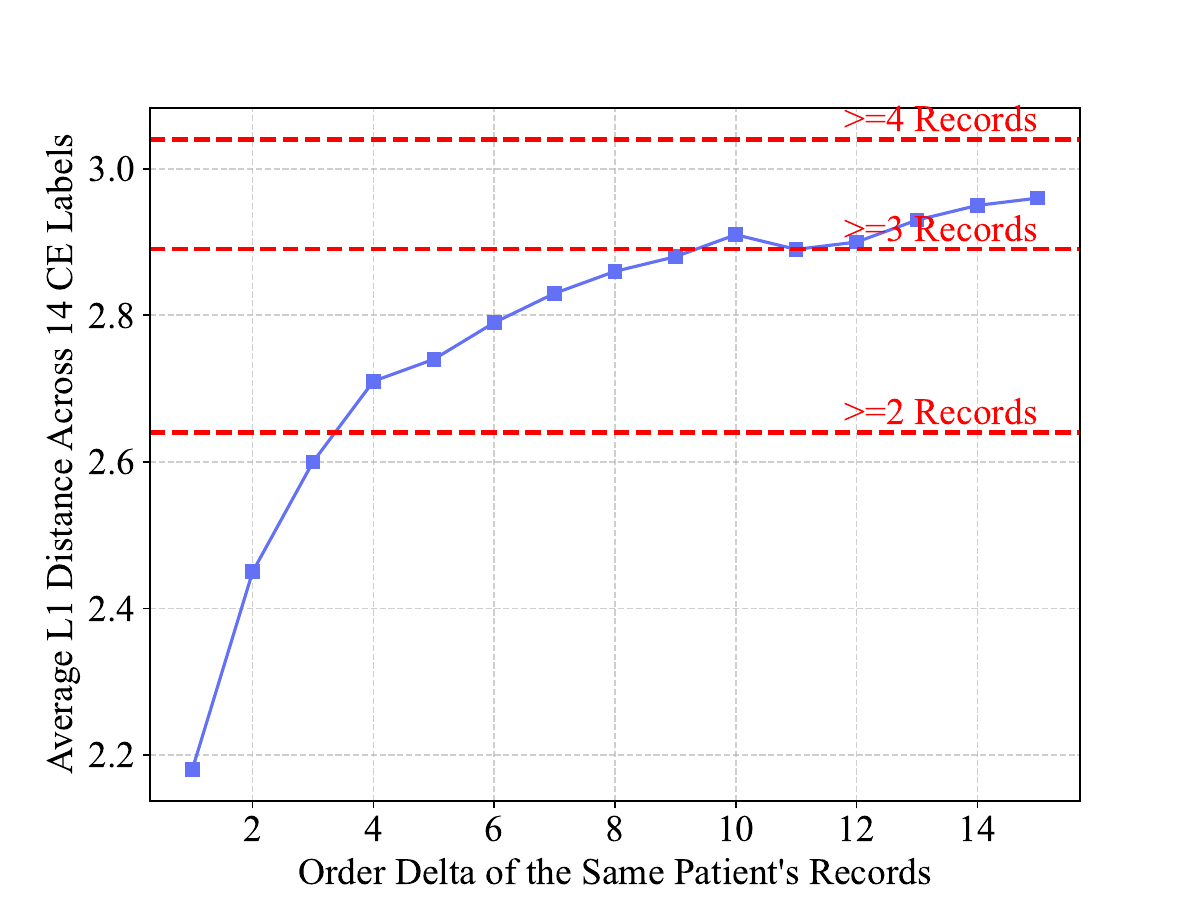}
        \label{fig:m-1}
    }
    \hspace{0.5cm}
    
    \caption{ 
    Figure (a) shows the distribution of the number of patients corresponding to different study counts. We can see that a large number of patients have only one study.
    Figure (b) shows the \textcolor[HTML]{6571EE}{blue line} representing the changes between two studies of the same patient in MIMIC-CXR as the study order increases. The \textcolor[HTML]{EA3323}{red dashed lines} indicate the differences across patients with a number of studies no less than specific values. 
    }
    \label{fig:tm}
\end{figure*}

\newpage
\section{Pilot Studies on Medical Diff-VQA}
\label{sec:plot_study}
\subsection{Motivation}
In the clinical diagnostic process, medical experts integrate a patient's medical historical records with current medical evidence to guide their diagnostic decisions. To explore whether this approach can enhance the diagnostic capabilities of large vision-language models in medical settings, we fine-tuned and evaluated InternVL2 8B and Qwen2-VL 8B on the publicly available Medical-Diff-VQA dataset~\cite{hu2023medical}, incorporating historical medical records.

\begin{figure}[ht]
    \centering\small
\begin{tcolorbox}[colframe=black]
\ttfamily
Assume you are a professional thoracic doctor. You are now provided with a chest radiology image. There is also a certain probability that you will be given the patient’s previous treatment image and report. Based on the provided information, please answer the corresponding questions. \\
The current study image is <image>.\\
The reference image is <image>.\\
The report of reference image is: \{historical records\}\\
The question is: \{question\}

\end{tcolorbox}
\caption{Text template on Medical-Diff-VQA using historical records.}
    \label{fig:medical-diff-vqa-with}
\end{figure}

\begin{figure}[ht]
    \centering\small
\begin{tcolorbox}[colframe=black]
\ttfamily
Assume you are a professional thoracic doctor. You are now provided with a chest radiology image and a question. Based on the provided information, please answer the corresponding questions.\\
The study image is <image>.\\
The question is: \{question\} 
\end{tcolorbox}
\caption{Text template on Medical-Diff-VQA without historical records.}
    \label{fig:medical-diff-vqa-without}
\end{figure}

The Medical-Diff-VQA dataset comprises seven categories of questions: abnormality, location, type, level, view, presence, and difference. However, we excluded the view questions, as they do not require historical information. Therefore, we selected six categories of questions from the Medical-Diff-VQA dataset for our study: \textbf{abnormality}, \textbf{location}, \textbf{type}, \textbf{level}, \textbf{presence}, and \textbf{difference}. The distribution of the training and test sets for each category is detailed in Table \ref{tab:number}. For each category, two distinct templates were used as model inputs: one incorporating historical records and the other devoid of additional information, as illustrated in Figure \ref{fig:medical-diff-vqa-with} and Figure \ref{fig:medical-diff-vqa-without}. These templates were employed to fine-tune and evaluate the model performance. It is worth noting that not all questions are associated with historical information; some are constructed based on records from a patient's initial visit, seen in Figure \ref{fig:m-3}. In such instances, we directly employed the template without additional information as the model input.

\begin{table*}[ht]
\centering\small
\begin{tabular}{l|cccccc}
\toprule
 & Abnormality & Level & Location & Presence & Type & Difference \\
\midrule
Training set & 116,394 & 53,728 & 67,187 & 124,654 & 22,067 & 131,563 \\
Test set & 14,515 & 6,846 & 8,496 & 15,523 & 2,702 & 16,389 \\
\bottomrule
\end{tabular}
\caption{The number of training and test samples for the six categories of questions in Medical-Diff-VQA.}
\label{tab:number}
\end{table*}

\begin{figure}[ht]
\includegraphics[width=0.5\textwidth]{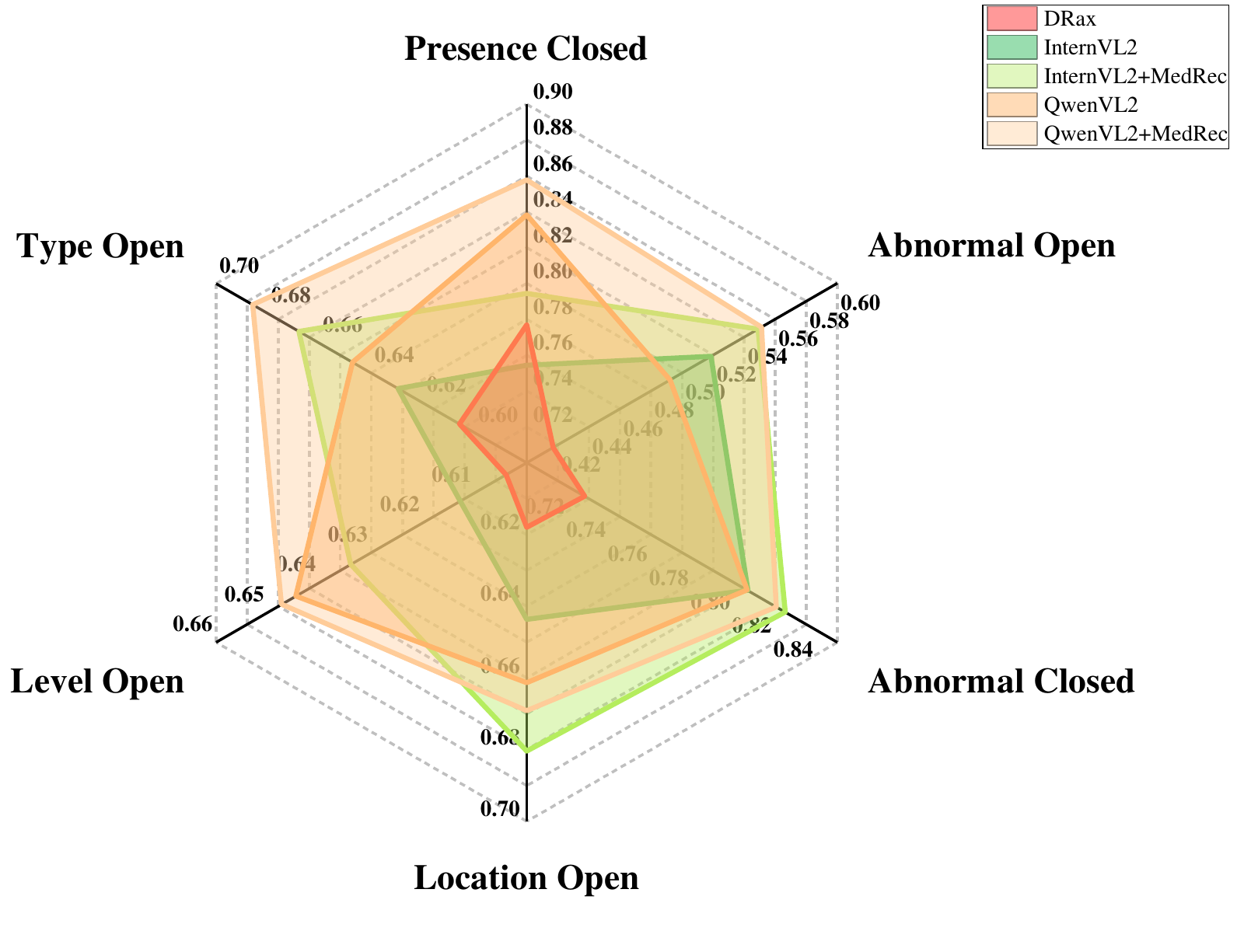}
\caption{Visualization of experimental results for different settings on the Medical-Diff-VQA dataset.}
\label{fig:m-2}
\end{figure}

For historical records, we utilized the report in MIMIC-CXR dataset, which contains data from 65,079 subjects, each representing an individual patient. A subject may have multiple studies, where each study corresponds to a patient's visit and includes multiple chest X-ray (CXR) images along with an associated medical report. As shown in Figure \ref{fig:m-1}, the blue line shows the average L1 distance variation between the Chexpert~\cite{ye2020weakly} labels of different studies within the same subject. The red dashed line indicates the average L1 distance between patients who meet certain conditions in terms of the number of studies. It is evident that as the study order increases, the differences between studies grow significantly. Furthermore, under the same conditions, the differences between patients are much greater than those within a single patient. Based on these observations, we \textbf{selected the most recent frontal CXR and report} as the historical records for our study.

\subsection{Experimental}
\textbf{Experimental Setting} We benchmarked our approach against the DRax method~\cite{10.1007/978-3-031-73471-7_10}, employing distinct evaluation metrics: \textbf{token recall} for open-ended questions and \textbf{accuracy} for close-ended ones. The models utilized in the experiments were InternVL2 8B and Qwen2-VL 8B. During the fine-tuning process, the LoRA rank was set to 16, the learning rate to 4e-5, and a cosine learning rate scheduling strategy was implemented. Training was conducted using the DeepSpeed Zero Stage 2~\cite{rajbhandari2020zero} distributed strategy on a 8-GPU RTX 3090 server.

\textbf{Experimental Results} Figure \ref{fig:m-2} shows the visualization of experimental results on 5 types of single-image questions (including closed-ended questions and open-ended questions). The results clearly indicate that integrating historical records as references significantly enhances VQA performance on both Qwen2-VL 8B and InternVL2 8B. 

\begin{table*}[ht!]
\centering
\resizebox{\textwidth}{!}{
\begin{tabular}{l|ccc|ccc|ccc|c}
\toprule
\multirow{2}{*}{\textbf{Method}} & \multicolumn{3}{c|}{\textbf{NLG metrics}} & \multicolumn{3}{c|}{\textbf{Missed Labels}} & \multicolumn{3}{c|}{\textbf{Added Labels}}& \multirow{2}{*}{\textbf{Average}}\\
\cmidrule{2-10}
 & \textbf{BLUE-1} & \textbf{Token Recall} & \textbf{Token F1} & \textbf{ML Acc} & \textbf{ML Recall} & \textbf{ML F1} & \textbf{AL Acc} & \textbf{AL Recall} & \textbf{AL F1} \\
\midrule
w/\ \ \  MedRecord& 0.464435 & 0.614229 & 0.59154 & 0.884127 & 0.267652 & 0.259081 & 0.885768 & 0.283076 & 0.275101 & 0.502779 \\
w/o MedRecord &\textbf{ 0.550117} & \textbf{0.684575} & \textbf{0.674029} & \textbf{0.947046} & \textbf{0.629887} & \textbf{0.615927} & \textbf{0.901559} & \textbf{0.345302} & \textbf{0.334627} & \textbf{0.631452} \\
\bottomrule
\end{tabular}
}
\caption{Evaluation results for the differences category in Medical-Diff-VQA using different methods.}
\label{tab:mdv6}
\end{table*}

\begin{figure*}[ht!]
\centering\small
\begin{tcolorbox}[colframe=black]
\ttfamily
\{\\
\hspace*{5mm}"qid": 53804,\\
\hspace*{5mm}"question": "Considering the changes observed, what can be concluded about the left lung in the recent CXR images?",\\
\hspace*{5mm}"question\_type": "single choice",\\
\hspace*{5mm}"content\_type": "worsen",\\
\hspace*{5mm}"options": \{\\
\hspace*{10mm}"A": "Worsened with increased effusion.",\\
\hspace*{10mm}"B": "Stable with no change.",\\
\hspace*{10mm}"C": "Improved or resolving condition.",\\
\hspace*{10mm}"D": "Complete resolution."\\
\hspace*{5mm}\},\\
\hspace*{5mm}"answer": "A",\\
\hspace*{5mm}"region\_name": "left lung",\\
\hspace*{5mm}"current\_image\_path": "files/p19/p19839145/s55247427/79b946ab-8d938e59-52027cad-8b4a4268-dab 951d0.jpg",\\
\hspace*{5mm}"regional\_current\_bbox": [1637, 122, 2796, 2196],\\
\hspace*{5mm}"history\_image\_path": "files/p19/p19839145/s59400960/36b4f554-95744e28-8b7baa81-fa621df8-41a d666f.jpg" ,\\
\hspace*{5mm}"regional\_history\_bbox": [1514, 463, 2442, 2278],\\
\hspace*{5mm}"regional\_history\_report": "Opacities in the left upper lobe have markedly improved. Moderate bilateral effusions larger on the left side with adjacent atelectasis are grossly unchanged. There is no pneumothorax.", \\
\hspace*{5mm}"global\_history\_report": "Mild cardiomegaly is stable. The aorta is tortuous. Cardiomediastinum is shifted to the right as before. Moderate bilateral effusions larger on the left side with adjacent atelectasis are grossly unchanged. Opacities in the left upper lobe have markedly improved. There is no pneumothorax."\\
\}
\end{tcolorbox}
\caption{A detailed data case sampled from the MMXU dataset.}
    \label{fig:detailed_data_case}
\end{figure*}

For the sixth category, the difference questions, we also investigated the effect of providing reference images along with historical records on the outcomes. These results are shown in Table \ref{tab:mdv6}. The responses to ``difference'' questions consist primarily of two components: the 14 ChexPert labels~\cite{ye2020weakly} that are either missing or added in the main image compared to the reference. The ``Missed Labels'' and ``Added Labels'' sections of the table employ accuracy, recall, and F1 scores for the quantitative assessment of the model outputs. Our findings indicate that the inclusion of historical records resulted in a substantial improvement across all evaluation metrics. (The 14 clinical efficacy labels in chexpert include: atelectasis, cardiomegaly, consolidation, edema, enlarged cardiomediastinum, fracture, lung lesion, lung opacity, no finding, pleural effusion, pleural other, pneumonia, pneumothorax, support devices.)

\begin{figure*}[ht!]
\centering\small
\begin{tcolorbox}[colframe=black]
\ttfamily
\{\\
\hspace*{5mm}"cmp\_id": 3,\\
\hspace*{5mm}"subject\_id": "14731346",\\
\hspace*{5mm}"study\_id": "54684841",\\
\hspace*{5mm}"cur\_image\_path": "files/p14/p14731346/s54684841/56ec2474-cfdbbdcd-5b133f15-6c0bc409-17435e c2.jpg",\\
\hspace*{5mm}"prior\_image\_path": "files/p14/p14731346/s51369333/33585d85-edf829eb-51303048-7bd9062a-b6ee3 a5b.jpg",\\
\hspace*{5mm}"cur\_report": "Right lower quadrant pain. In comparison with the study of \_\_\_, there has been worsening of the increased opacification at the left base with silhouetting of the hemidiaphragm and blunting of the costophrenic angle. These findings are consistent with a combination of volume loss in the left lower lobe and pleural effusion. Right lung is clear, and there is no evidence of vascular congestion.",\\
\hspace*{5mm}"prior\_report": "Hyperglycemia, intubated for airway protection, please assess NG tube and ET tube placement. AP radiograph of the chest was compared to prior study obtained the same day earlier. The ET tube tip is approximately 4.3 cm above the carina and slightly impinging the right wall of the trachea and should be repositioned. The NG tube tip is in the stomach. Right subclavian line tip is at the level of cavoatrial junction. Heart size and mediastinum are unremarkable. There is no pneumothorax. Minimal bibasilar opacities most likely reflect areas of atelectasis. No pulmonary edema is seen.",\\
\hspace*{5mm}"comp\_sent": "FINDINGS: In comparison with the study of \_\_\_, there has been worsening of the increased opacification at the left base with silhouetting of the hemidiaphragm and blunting of the costophrenic angle.",\\
\hspace*{5mm}"relationships": \{\\
\hspace*{10mm}"left lung": ["comparison|yes|worsened"],\\
\hspace*{10mm}"left lower lung zone": ["comparison|yes|worsened"],\\
\hspace*{10mm}"left costophrenic angle": ["comparison|yes|worsened"],\\
\hspace*{10mm}"left hemidiaphragm": ["comparison|yes|worsened"]\\
\hspace*{5mm}\},\\
\hspace*{5mm}"related\_region\_names": ["left lung", "left lower lung zone", "left costophrenic angle", "left hemidiaphragm"],\\
\hspace*{5mm}"cur\_image\_size": [2544, 3056],\\
\hspace*{5mm}"cur\_image\_bboxs": \{\\
\hspace*{10mm}"left lung": [1186, 777, 2087, 2210],\\
\hspace*{10mm}"left lower lung zone": [1241, 1705, 2087, 2210],\\
\hspace*{10mm}"left costophrenic angle": [1719, 1719, 1991, 1991],\\
\hspace*{10mm}"left hemidiaphragm": [1186, 1691, 2605, 2769] \\ 
\hspace*{5mm}\},\\
\hspace*{5mm}"prior\_image\_size": [[2544, 3056]],\\
\hspace*{5mm}"prior\_image\_bboxs": \{\\
\hspace*{10mm}"left lung": [[1268, 327, 2537, 2878]],\\
\hspace*{10mm}"left lower lung zone": [[1487, 1705, 2537, 2878]],\\
\hspace*{10mm}"left costophrenic angle": [[]],\\
\hspace*{10mm}"left hemidiaphragm": [[]] \\ 
\hspace*{5mm}\}\\
\}
\end{tcolorbox}
\caption{A detailed data case about \textbf{comparative sentence} during dataset construction.}
\label{fig:detailed_comp_sentence}
\end{figure*}

However, the Medical-Diff-VQA dataset does not fully capture the nuanced relationship between these questions and the evolution of historical historical records. This shortcoming complicates the determination of whether the model is truly leveraging historical information or simply reiterating prior data. Such behavior falls short of the expectations for real-world clinical applications. As a result, \textbf{there is an urgent need for a dataset that seamlessly integrates and balances historical medical records}, thereby supporting their meaningful utilization in practical healthcare settings.

\section{Detailed Data Case}
Figure \ref{fig:detailed_data_case} illustrates a detailed example from the MMXU dataset we developed, which is specifically tailored to analyze changes between two historical records of a patient related to chest X-rays. Each sample includes a single-choice question focused on changes in the condition of a particular region, along with comprehensive records from the patient's two visits. For the initial visit, the dataset provides images, bounding boxes marking the relevant regions, region-level reports, and overall-level reports. The questions are divided into three categories: ``Worsened'', ``Improved'', and ``No Change'', based on the comparative statements within the raw report. This dataset serves as a valuable resource for advancing research in multi-image comparison, historical record integration, and the application of visual grounding. 

\section{Detailed of Dataset Construction}
In this section, we will introduce some additional information during the  construction process.

\begin{figure*}[ht!]
    \centering
\begin{tcolorbox}[colframe=black]
\ttfamily
\small
You are a chest x-ray assistant and you are presented with a sentence involving a comparison in the original and current reports of a patient's two visits before and after, as well as some additional related explanatory information about the sentence. Please generate up to 3 single choice questions about the changes in the images from the two visits, based primarily on the sentence involving the comparison and the additional information, with the aid of referencing the two reports. Format the output into JSON format.
\\
\hrule
\vspace{5pt}
\textbf{Here are some rules:}
\vspace{5pt}\\
(1) Your questions are generated primarily around the KEY COMPARISON SENTENCE IN THE CURRENT REPORT and related explanatory information, with the content of the two reports as a secondary aid. \\
(2) Your question should be whether the chest CXR images have changed in some way in the same area, including abnormality, disease, location, severity, etiology, etc.\\
(3) For each question, you should clearly identify the area in which the question is asked and provide the correct answer, along with an explanation of why that answer was chosen.\\
(4) Your question is intended to answer the case where only images of the two diagnoses are known. Therefore, please do not include the words report comparison in your question; what should be included is image comparison.\\
(5) The JSON in related explanatory information consists of a "region" and a "list of explanatory information". For example, "right lung": ["comparison|yes|improved"] indicates that the right lung region has improved, and "left lung": ["comparison|yes|worsen"] indicates that the left lung region has worsened.\\
(6) Avoid asking questions that do not involve a change in two pictures. Ensuring all the questions are as diverse as possible.
\\
\hrule
\vspace{5pt}
\textbf{Here is one example:}\\
\{...\}\\
\hrule
\vspace{5pt}
\textbf{Here is some information:}\\
\{comparison sentence\},\ \{explanatory information\},\ \{current report\},\ \{prior report\}
\end{tcolorbox}
\caption{Text prompt for generating multiple image-based question-answer pairs in MMXU using GPT-4o}
    \label{fig:gpt4oprompt}
\end{figure*}

\subsection{Comparative Sentences Extraction}
\label{app:stage1}
Figure \ref{fig:detailed_comp_sentence} presents a detailed example of a comparative sentence extracted during the dataset construction process. The example includes the comparative sentence itself, the specific regions it pertains to, and the two corresponding images. Additionally, the bounding boxes of the relevant regions within the images, along with the comparative relationship, are illustrated. This detailed data structure is crucial for building a robust dataset that facilitates accurate comparisons during QA generation.

\subsection{QA Pairs Generation}
\label{app:stage3}
The version of GPT-4o we utilized is gpt-4o-2024-08-06. We employ the prompt template shown in Figure \ref{fig:gpt4oprompt}, where the "explanatory information" is derived from the details associated with the \textit{relationship} label. Figure \ref{fig:detailedgpt4oprompt} illustrates a detailed example of the input prompt and output content for the question-answer pair model using GPT-4o. We provided six specific rules and supplied GPT-4o with comparative sentences, \textit{relationship} labels, reports from two visits, and the region name of interest. Notably, the answers generated by GPT-4o are not limited to simple responses regarding improvements, deteriorations, or stability. Instead, we emphasize capturing more detailed changes in the regions.

\begin{figure*}[ht!]
\centering\small
\begin{tcolorbox}[colframe=black]
\ttfamily
Image-1: <image>\\
Image-2: <image>\\
Here are two chest X-RAY images of the same patient. Previous chest X-RAY is Image-1. And current chest X-RAY is Image-2.
Your task is to evaluate the differences between the two images based on the provided report and question.\\
\\
The report of Image-1 about is: \\
\%s\\
\\
Question:\\
\%s\\
\\
Options:\\
A: \%s\\
B: \%s\\
C: \%s\\
D: \%s\\
\\
Answer with the option's letter from the given choices directly.
\end{tcolorbox}
\caption{Text prompt template for evaluation on MMXU-\textit{test} benchmark with full report as historical records}
    \label{fig:templatewithfull}
\end{figure*}

\begin{figure*}[ht!]
\centering\small
\begin{tcolorbox}[colframe=black]
\ttfamily
Image-1: <image>\\
Image-2: <image>\\
Here are two chest X-RAY images of the same patient. Previous chest X-RAY is Image-1. And current chest X-RAY is Image-2.
Your task is to evaluate the differences between the two images based on the provided regional report and question.\\
\\
The report of Image-1 about region \%s is: \\
\%s\\
\\
Question:\\
\%s\\
\\
Options:\\
A: \%s\\
B: \%s\\
C: \%s\\
D: \%s\\
\\
Answer with the option's letter from the given choices directly.
\end{tcolorbox}
\caption{Text prompt template for evaluation on MMXU-\textit{test} benchmark with regional report as historical records}
    \label{fig:templatewithregional}
\end{figure*}

\section{Evaluation on MMXU-\textit{test} benchmark}

\noindent\textbf{Experimental Settings} We respectively utilized specialized prompt templates for three distinct scenarios: direct evaluation, enhancement through the integration of global and regional historical reports as historical records. For all generation processes, we set  {do\_sample} to  {False} and  {temperature} to  {0}. During the InternVL2 8B fine-tuning process, we set  {epochs} to  {1},  {max\_dynamic\_patch} to  {6},  {down\_sample\_ratio} to  {0.5},  {lora\_rank} to  {16}, and the  {learning\_rate} to  {4e-5}. The learning rate scheduler type is set to "cosine," with a warmup ratio of  {0.03}. The fine-tuning is carried out using the DeepSpeed Zero Stage 2~\cite{rajbhandari2020zero} distributed strategy on an 8-GPU NVIDIA RTX 3090 server. 

\begin{figure*}[ht]
\centering\small
\begin{tcolorbox}[colframe=black]
\ttfamily
Image-1: <image>\\
Image-2: <image>\\
Here are two chest X-RAY images of the same patient. Previous chest X-RAY is Image-1. And current chest X-RAY is Image-2.
Your task is to evaluate the differences between the two images based on the provided question.\\
\\
Question:\\
\%s \\
\\
Options:\\
A: \%s\\
B: \%s\\
C: \%s\\
D: \%s\\
\\
Answer with the option's letter from the given choices directly.
\end{tcolorbox}
\caption{Text prompt template for direct evaluation on MMXU-\textit{test} benchmark without historical records}
    \label{fig:templatewithout}
\end{figure*}

\noindent\textbf{Text Prompt} To assess the impact of historical  records on medical diagnosis, we performed evaluations on our MMXU-\textit{test} dataset under three distinct conditions: without historical records, with global reports as historical records, and with regional reports as historical records. Figure \ref{fig:templatewithout} illustrates the prompt template used in the absence of historical records, where only the basic instructions, questions, and options are provided. Figure \ref{fig:templatewithfull} depicts the prompt template incorporating global reports as historical records, in which we enhance the original template by including the global report. Figure \ref{fig:templatewithregional} showcases the prompt template with regional reports as historical records. In contrast to the global reports, these regional reports are tailored to the specific region relevant to the question, with the region's name explicitly indicated.

When generating region-level reports as historical record enhancements, we distinguish regions using the 29 anatomical names provided in the Chest ImaGenome dataset, including: right lung, right upper lung zone, right mid lung zone, right lower lung zone, right hilar structures, right apical zone, right costophrenic angle, right hemidiaphragm, left lung, left upper lung zone, left mid lung zone, left lower lung zone, left hilar structures, left apical zone, left costophrenic angle, left hemidiaphragm, trachea, spine, right clavicle, left clavicle, aortic arch, mediastinum, upper mediastinum, svc, cardiac silhouette, cavoatrial junction, right atrium, carina, and abdomen.

\begin{figure*}[ht]
  \centering
  \includegraphics[width=2.05\columnwidth]{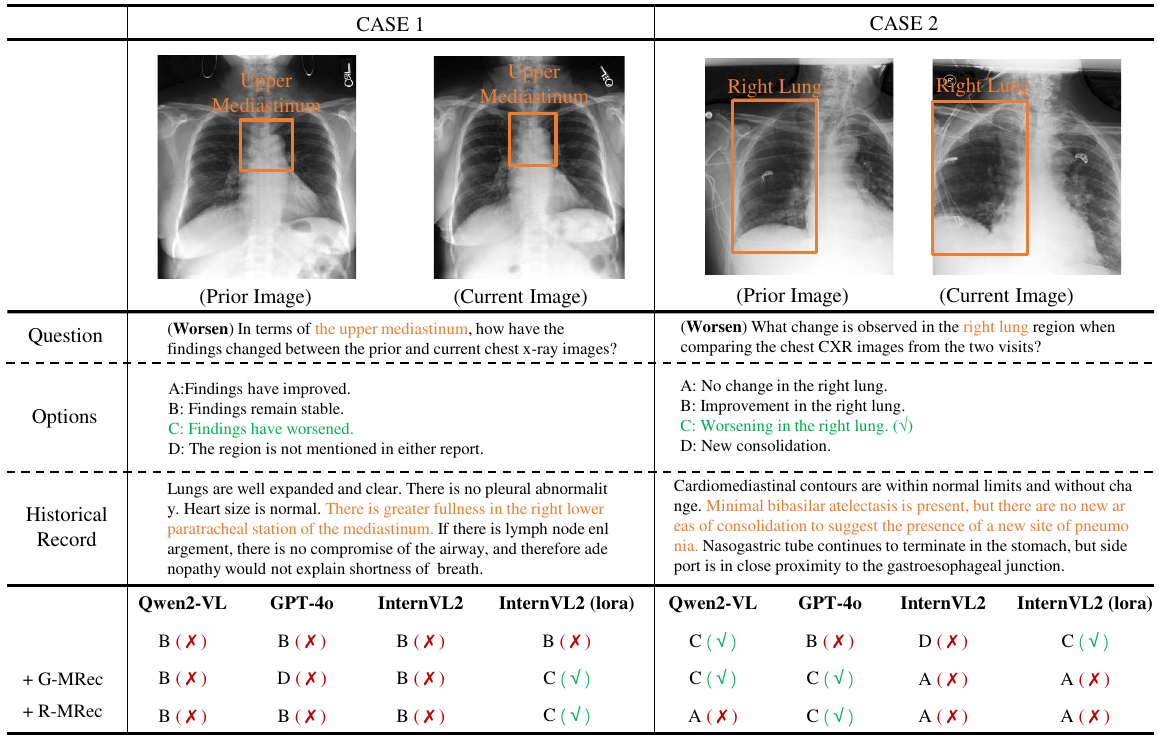}
  \caption{Two hard examples from ``Worsen'' category of MMXU answered by some LVLMs. }
  \label{fig:case-worsen}
\end{figure*}

\section{More Case Study}
\label{app:more-case-study}


Figure \ref{fig:case-worsen} shows two challenging examples from the "Worsen" category in MMXU. In CASE 1, none of the models without fine-tuning were able to correctly answer the question, and only the fine-tuned model using the MAG method provided the correct answer. This may be due to the difficulty of the question itself, as well as the challenge of extracting useful information to answer the question from the historical records.In CASE 2, only GPT-4o was able to answer the question correctly after providing historical records. For other models, since R-MRec reiterated previously unchanged views, all incorrectly chose option A (no change), indicating that these models failed to correctly understand the change and simply repeated the previous information.

\begin{figure*}[ht]
    \centering

\resizebox{0.98\textwidth}{!}{
\begin{tcolorbox}[colframe=black]
\ttfamily
\small
\textbf{USER:} You are a chest x-ray assistant and you are presented with a sentence involving a comparison in the original and current reports of a patient's two visits before and after, as well as some additional related explanatory information about the sentence. Please generate at most 3 single choice questions about the changes in the images from the two visits, based primarily on the sentence involving the comparison and the additional information, with the aid of referencing the two reports. Format the output into JSON format.
\\
Here are some rules:
\\
(1) Your questions are generated primarily around the KEY COMPARISON SENTENCE IN THE CURRENT REPORT and related explanatory information, with the content of the two reports as a secondary aid. \\
(2) Your question should be whether the chest CXR images have changed in some way in the same area, including abnormality, disease, location, severity, etiology, etc.\\
(3) For each question, you should clearly identify the area in which the question is asked and provide the correct answer, along with an explanation of why that answer was chosen.\\
(4) Your question is intended to answer the case where only images of the two diagnoses are known. Therefore, please do not include the words report comparison in your question; what should be included is image comparison.\\
(5) The JSON in related explanatory information consists of a "region" and a "list of explanatory information". For example, "right lung": ["comparison|yes|improved"] indicates that the right lung region has improved, and "left lung": ["comparison|yes|worsen"] indicates that the left lung region has worsened.\\
(6) Avoid asking questions that do not involve a change in two pictures. Ensuring all the questions are as diverse as possible.
\\
Here is one example:\\
\{ \\
\hspace*{5mm}"question": ..., \\
\hspace*{5mm}"region": ...,\\
\hspace*{5mm}"content\_type": ..., \# worsen, improved or no change\\
\hspace*{5mm}"options": \{ \\
\hspace*{10mm}"A": ..., \\
\hspace*{10mm}"B": ..., \\
\hspace*{10mm}"C": ..., \\
\hspace*{10mm}"D": ... \}\\
\hspace*{5mm}"answer": .., \\
\hspace*{5mm}"reason": ...\\
\}\\
Here is some information: \\
KEY COMPARISON SENTENCE IN THE CURRENT REPORT: "Widening mediastinum has improved." \\
Explanatory Information: "upper mediastinum": ["comparison|yes|improved"] \\
Current Report: Moderate to severe cardiomegaly is stable.  Widening mediastinum has improved. Bilateral effusions are small. Opacities in the lower lobes bilaterally larger on the left side are likely atelectasis, superimposed infection cannot be totally excluded. There is no evident pneumothorax. Sternal wires are aligned. Degenerative changes in the thoracic spine are mild.  There is mild vascular congestion. Bilateral healed rib fractures are again noted. Small bilateral effusions with adjacent opacities, these are likely atelectasis but superimposed infection cannot be totally excluded. Mild 
vascular congestion.\\
Prior Report: IN COMPARISON TO \_\_\_ RADIOGRAPH, CARDIOMEGALY IS ACCOMPANIED BY IMPROVING PULMONARY EDEMA.  A MORE CONFLUENT OPACITY IN THE LEFT RETROCARDIAC REGION HAS WORSENED, AND MAY REFLECT A COMBINATION OF PLEURAL effusion and adjacent atelectasis or consolidation.  No other relevant change.
\vspace{5pt}
\hrule
\vspace{5pt}
\textbf{GPT-4o:} \\
\{\\
\hspace*{5mm}"question": "What change has been observed in the upper mediastinum region of the chest CXR images?", \\
\hspace*{5mm}"region": "upper mediastinum",\\
\hspace*{5mm}"content\_type": "improved",\\ 
\hspace*{5mm}"options": \{ \\
\hspace*{10mm}"A": "The widening has worsened.", \\
\hspace*{10mm}"B": "There is new vasculature congestion.", \\
\hspace*{10mm}"C": "The widening has improved.", \\
\hspace*{10mm}"D": "There is no notable change."\\
\hspace*{5mm}\}, \\
\hspace*{5mm}"answer": "C",  \\
\hspace*{5mm}"reason": "The related explanatory information indicates that there is an improvement in the upper mediastinum, corresponding to the general improvement of the mediastinum's widening mentioned in the key comparison sentence."\\
\}

\end{tcolorbox}
}
\caption{A detailed example of the input and output content for the question-answer pair model using GPT-4o}
\label{fig:detailedgpt4oprompt}
\end{figure*}

\end{document}